\documentclass[10pt,twocolumn,letterpaper]{article}

\usepackage[pagenumbers]{cvpr} 

\usepackage[dvipsnames]{xcolor}

\usepackage{booktabs} 
\usepackage{colortbl}
\usepackage{svg}
\usepackage{lipsum}
\usepackage{multirow}

\usepackage[ruled,vlined]{algorithm2e}
\SetAlgoHangIndent{2cm}

\definecolor{commentcolor}{RGB}{110,154,155} 
\newcommand{\PyComment}[1]{\ttfamily\textbf{\textcolor{commentcolor}{\# #1}}}
\newcommand{\PyCode}[1]{\ttfamily\textbf{\textcolor{black}{#1}}}

\usepackage[title]{appendix}

\usepackage{tabularx}

\definecolor{cvprblue}{rgb}{0.21,0.49,0.74}
\usepackage[pagebackref,breaklinks,colorlinks,citecolor=cvprblue]{hyperref}

\def\paperID{14232} 
\def\confName{CVPR}
\def\confYear{2024}

\title{Diffusion-Based Particle-DETR for BEV Perception}

\author{
Asen Nachkov$^{1}$ \quad Martin Danelljan$^{2}$ \quad Danda Pani Paudel$^{1, 2}$ \quad Luc Van Gool$^{1, 2}$ \\
\\
$^{1}$INSAIT, Sofia University \qquad $^{2}$ETH Zurich
}

\begin{document}
\maketitle

\begin{abstract}
    The Bird-Eye-View (BEV) is one of the most widely-used scene representations for visual perception in Autonomous Vehicles (AVs) due to its well suited compatibility to downstream tasks. For the enhanced safety of AVs, modeling perception uncertainty in BEV is crucial. Recent diffusion-based methods offer a promising approach to uncertainty modeling for visual perception but fail to effectively detect small objects in the large coverage of the BEV. Such degradation of performance can be attributed primarily to the specific network architectures and the matching strategy used when training. Here, we address this problem by combining the diffusion paradigm with current state-of-the-art 3D object detectors in BEV. We analyze the unique challenges of this approach, which do not exist with deterministic detectors, and present a simple technique based on object query interpolation that allows the model to learn positional dependencies even in the presence of the diffusion noise. Based on this, we present a diffusion-based DETR model for object detection that bears similarities to particle methods. Abundant experimentation on the NuScenes dataset shows equal or better performance for our generative approach, compared to deterministic state-of-the-art methods. Our source code will be made publicly available.
\end{abstract}    
\section{Introduction}
\label{sec: introduction}

3D Object detection - the task of localizing and classifying objects in a real-world 3D coordinate frame - is one of the most important tasks in the pipeline of an autonomous vehicle. It is critical to safe self-driving since it informs the subsequent prediction, planning, and actuation modules and, evidently, one needs to recognize an obstacle in order to avoid it. Estimating the object locations directly from the camera views \cite{brazil2019m3d, brazil2020kinematic, mao20233d} faces difficulties related to perspective warping and size-distance ambiguities. Instead, the bird-eye-view (BEV) has established itself as a useful representation to facilitate perception because it is ego-centered, metrically-accurate, orthographic - thus avoiding perspective distortion of shapes, and suffers less from occlusions and object deformations.

Recently, it has been shown that diffusion models can be successfully used for 2D object detection \cite{chen2023diffusion} - a completely different setup than the generative tasks like text-to-image where they have been dominating \cite{rombach2022high, song2020denoising, ho2020denoising, croitoru2023diffusion, saharia2022photorealistic}. In principle, one should then be able to apply diffusion-based object detection also in the 2D BEV and predict 3D locations, reaping all the diffusion benefits like incremental refinement and the ability to trade-off compute and accuracy?

\begin{figure}[tbp]
    \centering
    \includegraphics[width=\linewidth]{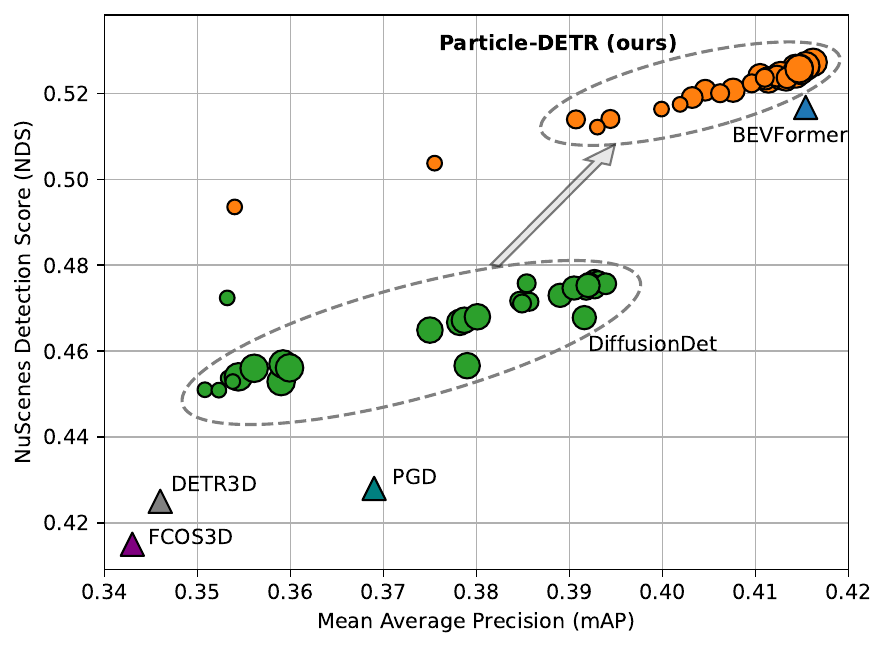}
    \captionsetup{aboveskip=0cm, belowskip=-0.4cm}
    \caption{\textbf{Performance improvement}. Our method outperforms the baseline stochastic model DiffusionDet \cite{chen2023diffusion} and is comparable in performance to deterministic models like BEVFormer \cite{li2022bevformer}. Here, the results of stochastic models are shown as circles and those of deterministic models as triangles. The size of the circles is proportional to the number of search tokens used.}
    \label{fig: contributions}
\end{figure}

We find that naively applying diffusion in BEV yields an algorithm with insufficient performance. We attribute this to the challenging problem setting and the fact that the network architecture is not tuned for the particular geometric aspects of the BEV:
\begin{enumerate}
    \item \textbf{Setting}: Recent works \cite{li2022delving, philion2020lift, li2022bevformer, roddick2018orthographic} represent the BEV as a set of spatially-correlated latent features corresponding to a $(50 \times 50)$ or even $(100 \times 100)$ meter grid around the ego-vehicle. The detectable objects such as cars and pedestrians are naturally very small in relation to the size of the whole BEV map, making detection more challenging compared to on common datasets used to benchmark 2D detection algorithms \cite{lin2014microsoft, deng2009imagenet, everingham2010pascal}.
    \item \textbf{Architecture}: DiffusionDet \cite{chen2023diffusion}, the representative diffusion algorithm, uses an ROI-based architecture \cite{girshick2014rich} which aggregates BEV features only within the proposed boxes. This makes object features rather local, preventing extensive search on the BEV. Local features work well in settings where the target boxes are larger and more dense, but we believe in the BEV one needs a more specialized architecture to better handle object sparsity. 
\end{enumerate}

\paragraph{Problem statement and approach.} Since object detection is ultimately a search problem and smaller objects are harder to locate, some of the inherent challenges when using diffusion to detect objects can be exacerbated in the BEV. Thus, the research question we try to answer is: \emph{How should the diffusion approach and network architecture be adjusted so as to ease the search process in the BEV?} To that end, our insights are that first, to search more effectively, one should pool information across the \emph{search tokens} used (boxes, anchors, queries), and second, one should take measures to prevent the diffusion noise from overwhelming any positional dependencies that exist in the data.

To pool information across the search tokens we need to have them \emph{communicate} with each other. This can be achieved using self-attention which in turn points to a transformer method like DETR \cite{carion2020end, zhu2020deformable, zhang2022dino, liu2022dab, li2022dn, dai2021dynamic}. These models utilize a number of fixed \emph{object queries} which they learn to regress into the predicted boxes. They further make incorporate cross-attention modules to look up relevant features from the image in a way that is independent across individual queries. The combined architecture can utilize global features, which becomes increasingly more useful as the objects' sizes decrease. 

Regarding positional dependencies, we show how the diffusion noise affects the matchings between predicted and target boxes. In essence, most approaches like Deformable-DETR \cite{zhu2020deformable} start with fixed $(x, y)$ reference points, look up the image features in those locations, and output corrections which are subsequently applied to them. But when diffusion is applied on the initial reference points, they become no longer associated with the object queries, preventing the model from using positional information. To address this challenge we introduce \emph{object query interpolation} as a simple way to learn positional relations for DETR-like models even in the presence of noise over the references.

The resulting generative model can refine its predictions, trade-off accuracy and compute, operate with a different number of search tokens at train and test time, and yields results comparable or better than those of battle-tested deterministic models. Furthermore, it has similarities to particle methods from which ideas like particle pruning and refinement can be borrowed.

\paragraph{Contributions.} Our contributions are the following:
\begin{enumerate}
    \item In Section \ref{subsection: matching} we provide a novel view on the assignment instability problem that is present in most DETR-like models \cite{carion2020end, zhu2020deformable} by showcasing how the stochasticity of the diffusion process affects the assignments.
    \item In Section \ref{subsec: query_interpolation} we showcase our module called \emph{query interpolation} which allows the model to learn positional information even in the presence of diffusion noise.
    \item We integrate the proposed module into a deformable-DETR \cite{zhu2020deformable} variant, called Particle-DETR, which uses diffusion to denoise box centers to their true positions. We further provide a detailed analysis of the performance of the model on the realistic and large-scale NuScenes dataset \cite{caesar2020nuscenes}.
\end{enumerate}
\begin{figure}[tbp]
    \centering
    \includegraphics[width=\linewidth]{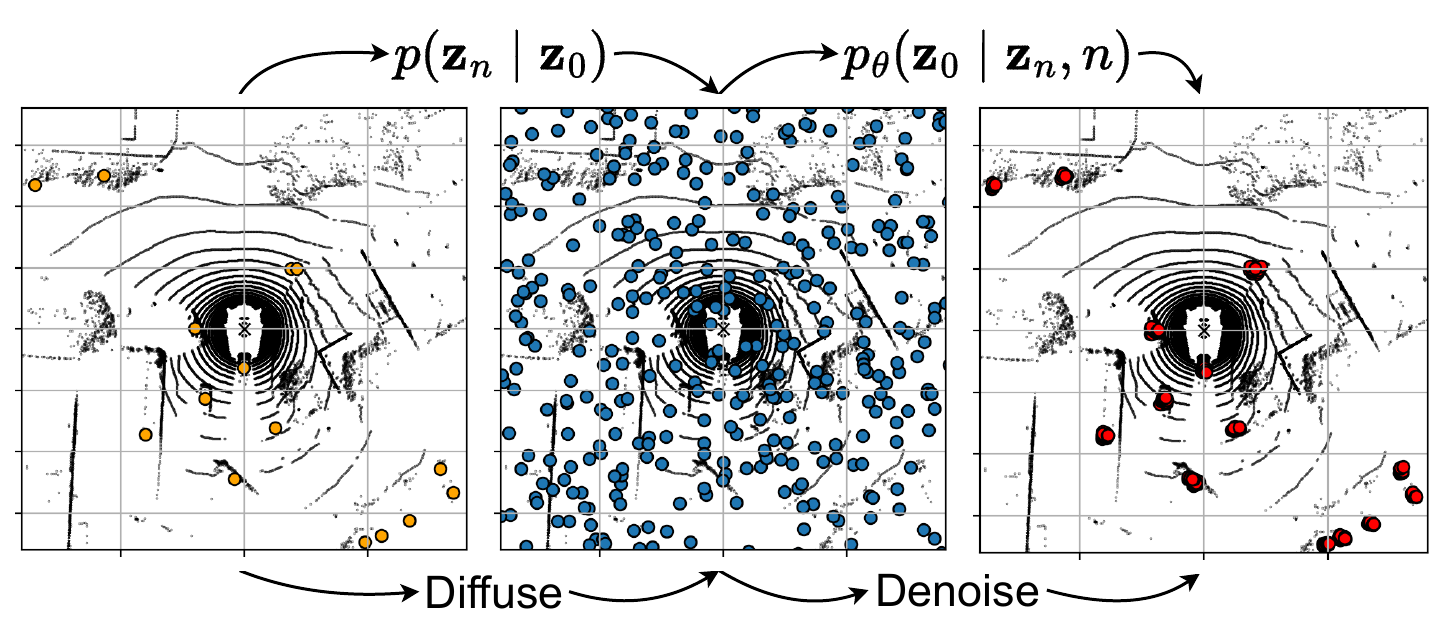}
    \captionsetup{aboveskip=0cm, belowskip=-0.4cm}
    \caption{\textbf{Diffusion in BEV}. Our approach diffuses the ground-truth object centers in BEV and learns to denoise them. At test time, we start from random references corresponding to the box centers and progressively refine them to their true locations.}
    \label{fig: high level process}
\end{figure}

\section{Related Work}
\label{sec: related_work}

\begin{figure*}[tbp]
    \centering
    \includegraphics[width=\linewidth]{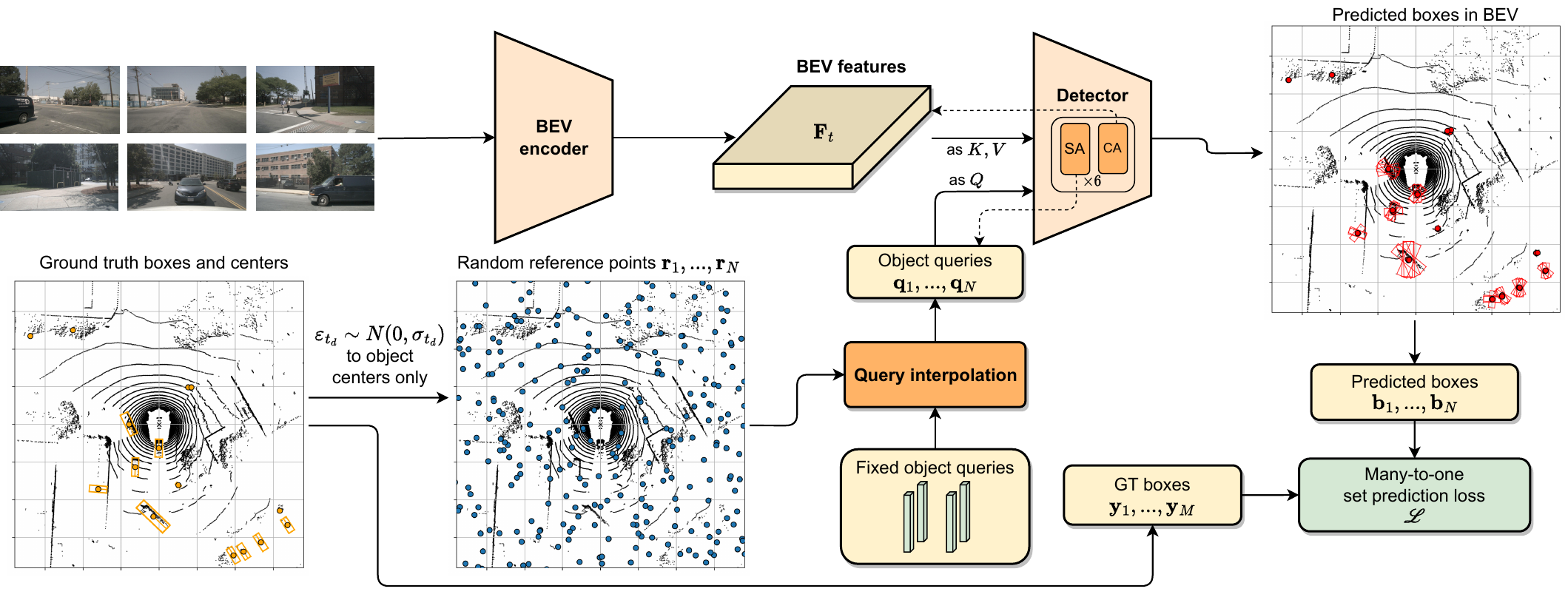}
    \caption{\textbf{Diffusion in BEV}. A feature extractor processes the images from the camera feed at the current timestamp. An encoder learns to project these features from the perspective of the car to the top-down orthographic bird-eye-view. At training time we add noise to the ground-truth object centers according to a diffusion schedule, while at test time we directly sample the references. Using our query interpolation module each reference position is associated with a spatial feature. Using these features, the decoder learns to denoise the reference points to their true locations. A set prediction loss is adopted for training.}
    \label{fig: high_level_overview}
\end{figure*}

\textbf{Diffusion-based object detection.} Utilizing a diffusion model for exact tasks started with DiffusionDet \cite{chen2023diffusiondet}, where the model learns to denoise axis-aligned 2D boxes in the image coordinate system. First, a backbone network, e.g., ResNet \cite{he2016deep} or Swin transformer \cite{liu2021swin}, extracts multi-scale image features. At train time, random noise is added to the ground-truth (GT) boxes according to a diffusion schedule, while at test time random boxes are sampled from a Gaussian distribution. Subsequently, a decoder with a region-of-interest-based (ROI) architecture \cite{he2017mask, girshick2014rich} aggregates the features inside each box and produces corrections to the box parameters. The output boxes are then matched to the GT boxes for training.

\textbf{Other applications.} Inspired by DiffusionDet, the usage of diffusion models for other prediction tasks has increased. It has been applied to the denoising of BEV features \cite{zou2023diffbev}, to prediction of future discrete BEV tokens \cite{zhang2023learning}, to action segmentation in videos \cite{liu2023diffusion}, to weakly supervised object localisation (WSOL) \cite{zhao2023generative}, to human motion prediction \cite{barquero2023belfusion} and pose estimation \cite{feng2023diffpose}, to domain adaptive semantic segmentation \cite{peng2023diffusion}, to video anomaly detection \cite{tur2023exploring}, to camouflaged object detection \cite{chen2023diffusion}, to text-video retrieval \cite{jin2023diffusionret}, and to open-world object detection \cite{wang2023random}.
\\ 

\textbf{The DETR family of models.} Current object detection in BEV is dominated by DETR-variants \cite{carion2020end, liu2022dab, li2022dn, dai2021dynamic, zhang2022dino, chen2023group, ye2023cascade}. They utilize a transformer sequence where a fixed number of object queries look up the relevant image features using cross-attention and are transformed to a fixed number of output boxes. Since the outputs and the GTs are unordered, a set-matching step is needed in order to assign predictions to targets. This matching has been described as \emph{unstable}, because of how one prediction can be matched to different targets across multiple training iterations on the same image. Various approaches try to mitigate this issue by introducing query denoising \cite{li2022dn}, where some object queries are matched to their target by index, and contrastive denoising \cite{zhang2022dino} where both positive and negative examples are used in each query group.

\textbf{BEV perception.} Transforming the camera features to BEV is an active area of research. It has been done using both traditional approaches where 3D voxels are projected onto the image plane and the image features within the projection are average-pooled \cite{roddick2018orthographic}, or where a categorical depth distribution is estimated for each image pixel, after which the features are "lifted" in 3D according to their depths \cite{philion2020lift, reading2021categorical}. Implicit projection, where depth is not estimated explicitly, can be achieved by utilizing self-attention to look up the past BEV and cross-attention to look up the current image features \cite{li2022bevformer, yang2023bevformer, qin2023unifusion, zhang2022beverse, jiang2023polarformer, huang2022bevdet4d}. This is the approach we rely on in this work. Once in BEV, models may perform joint detection and trajectory prediction \cite{dendorfer2022quo, hu2022st, hu2021fiery, zhang2022beverse, wu2020motionnet, luo2018fast}, BEV segmentation \cite{peng2023bevsegformer}, tracking \cite{gwak2022minkowski}, or agent interaction analysis \cite{cui2023gorela, chai2019multipath}.
\section{Approach}
\label{sec: approach}

In this section we motivate our method by considering the unique challenge arising when combining diffusion with perception in BEV, cf. Subsection \ref{subsec: conceptual} and how our method alleviates this challenge, cf. Subsection \ref{subsec: query_interpolation}.

\subsection{Preliminaries}
\paragraph{Diffusion models.} Diffusion models are a class of generative models whose goal is to learn to sample from the distribution over a sample space. To that end, as part of the training procedure, a stochastic process adds noise to each input sample according to a predefined schedule. At training time, the model learns to predict the added noise, while at test time one generates random initial noise which the model progressively denoises until a data point from the training distribution is formed.

The \emph{forward} process, which adds noise to the sample at training time, is defined as 

$$
q(\mathbf{z}_{t_d} | \mathbf{z}_0) = \mathcal{N}(\mathbf{z}_{t_d} | \sqrt{\bar{\alpha}_{t_d} \mathbf{ z}_0}, (1 - \bar{\alpha}_{t_d}) \mathbf{I}),
$$
where $t_d$ is the time-index of the diffusion process (different from the temporal frame index $t_f$ in the context of the BEV sequences), $\mathbf{z}_{t_d}$ is the noisy sample at that time, $\mathbf{z}_0$ is the noise-less ground-truth sample, and $\bar{\alpha}_{t_d} = \prod_{s = 0}^{t_d} \alpha_s = \prod_{s=0}^{t_d} (1 - \beta_s)$ is the corresponding parameter from the schedule controlling the variance of the noise.

The network output $f_\theta(\mathbf{z}_{t_d}, t_d)$ is conditioned on the noisy sample $\mathbf{z}_{t_d}$, the diffusion time $t_d$ and its parameters $\theta$ are optimized to minimize the loss 

$$
\mathcal{L} = \frac{1}{2}{\lVert f_\theta (\mathbf{z}_{t_d}, t_d) - \mathbf{z}_0 \rVert}^2.
$$

Since this corresponds to a denoising process, at test time we sample random noise $\mathbf{z}_T$ and progressively refine it by feeding the previous output as the subsequent input to the network, i.e. $\mathbf{z}_0 = f_\theta( f_\theta(...(f_\theta( f_\theta (\mathbf{z}_T, T), T-1)) ...), 0) $. Various improvements exist to speed-up this process \cite{mao2023leapfrog, song2020denoising, nichol2021improved, chung2022come}.

Since the noise added to each data sample is independent across all sample elements, we can use this process to generate different objects like images \cite{rombach2022high}, bounding boxes \cite{chen2023diffusiondet}, camera poses \cite{wang2023posediffusion}. Here, the diffusion is applied over box centers $(c_x, c_y)$ in BEV, to which we refer as \emph{particles}.

\textbf{DETR models.}
DETR models for object detection \cite{carion2020end} rely on a transformer-based architecture. A feature extractor, usually convolutional, extracts image features which are then passed to a transformer encoder where each feature patch can attend to other feature patches. Subsequently, a transformer decoder, relying on a fixed number $N$ of latent vectors $\{{ \mathbf{q}_1, ..., \mathbf{q}_N \}} $ called \emph{object queries}, looks up the features from the encoder and outputs bounding boxes. A one-to-one matching step using the Hungarian algorithm is required to assign predictions to box targets.

The object queries are learned using gradient descent and are fixed at test time. Since positional encodings for the object queries are also used, the model can learn information related to the order of the object queries.

An important modification to the this setup is given by DeformableDETR \cite{zhu2020deformable} where $\{{ \mathbf{q}_1, ..., \mathbf{q}_N \}} $ are not only ordered between them, but each object query $\mathbf{q_i}$ is tied to a particular 2D position $\mathbf{r}_i$, called a reference point within the image coordinate frame. Since both the object queries and reference points are learned, the model can focus not only on the content of the pixels, but also on the query positions.

Fixing the reference points $\{{\mathbf{r}_1, ..., \mathbf{r}_N \}} $ makes training easier because query $\mathbf{q}_i$ will always have the same relative location compared to query $\mathbf{q}_j, j \ne i$. In that case, the cross-attentions in the decoder learn only how to attend to the surrounding features which makes learning more stable. This fixed nature is crucial in relation to the stochasticity we will introduce by the diffusion process.

\subsection{Adding diffusion to BEV}
\label{subsec: conceptual}

\begin{figure}[tbp]
    \centering
    \includegraphics[width=\linewidth]{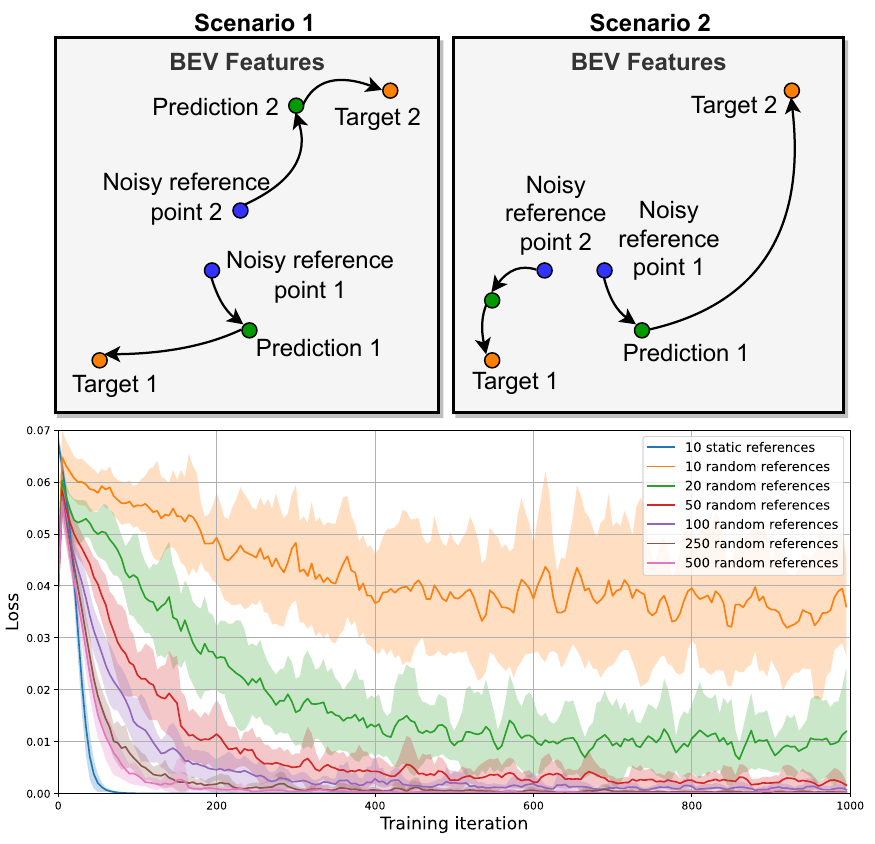}
    \captionsetup{aboveskip=0cm, belowskip=-0.4cm}
    \caption{\textbf{Label ambiguity}. A case where the random reference points may produce different targets, depending on the matching. The top row shows ambiguity when we match predictions to GTs by total distance (linear sum assignment). The columns show how different samplings of the blue points may push the same feature location in BEV to different targets, thus confusing the model. In all cases the matching is done between the predictions and the targets. The bottom row shows how the training loss depends on the number of references for a simple toy task (cf. suppl. materials).}
    \label{fig: label_ambiguity}
\end{figure}

Our setup is shown in Figure \ref{fig: high_level_overview}. A feature extractor \cite{he2016deep} along with a feature neck \cite{lin2017feature} processes all camera images from the current timestep $t_f$, outputting multi-scale feature maps for each camera view. A BEV encoder, in practice BEVFormer \cite{li2022bevformer}, projects these features around the ego-vehicle. In BEV, we add noise to the ground-truth object centers and concatenate with additional random locations. These are passed as reference points to the decoder which, similar to DeformableDETR \cite{zhu2020deformable}, refines some of them into the GT positions.

At test time, we sample initial random box centers and run them through the decoder. Since the model is trained to work with variable reference points, it can plug in the predicted box centers as input reference points in the next denoising step. This allows iterative refinement of our predictions - something that deterministic models like DeformableDETR \cite{zhu2020deformable} cannot do because they rely on object queries which are fixed to particular positions.

We follow DETR \cite{carion2020end} in applying auxiliary losses to each decoder layer, instead of just the final one. We refer to each decoder layer as a \emph{stage} and to each pass through all decoder layers as a single DDIM \cite{song2020denoising} step. By having multiple such steps we can trade-off accuracy and compute. Each DDIM step requires evaluating the only the decoder.

\subsection{Matching}
\label{subsection: matching}

The matching cost used in object detectors from the DETR \cite{carion2020end} family typically considers both the predicted box dimensions \emph{and} the predicted class logits. As a result, one cannot say that predictions spatially closer to the GT box will \emph{always} be matched and those farther away will not. Yet, deterministic detectors do converge because even if the matching changes across iterations, the static nature of the inputs allows one to learn the spatial relationships in the image.

\paragraph{Label ambiguity.} In the diffusion case there exist specific situations where learning is, in fact, \emph{impossible} due to the same BEV feature having different targets, depending on the noisy sampling of the reference points. Figure \ref{fig: label_ambiguity} illustrates these conceptually.

Suppose we use the Hungarian algorithm for matching and we sample the initial reference points as the blue points on the top-left plot in Figure \ref{fig: label_ambiguity}. Then the matching will be as shown by the arrows. However, if one of the points is sampled differently, as in the top-right plot, we may end up matching them differently. In reality, the BEV features corresponding to the $(x, y)$ position where noisy reference point 1 is, will be pushed by the optimization in the first case toward target 1 and in the second case toward target 2. This creates label ambiguity arising \emph{specifically} due to the random sampling of the starting locations.

Using more object queries than GT boxes reduces the possibility of this ambiguity to hinder the training. This is because having more predictions and matching them with any strategy that takes the distance into account (unlike e.g. matching by index) will make the model produce smaller refinements to the starting reference points. Thus, a high amount of object queries is needed both to detect many objects, but also to help detect them accurately. Explanations on a toy example can be found in Section \ref{supp: label_ambiguity} from the supplementary materials.

\subsection{Object query interpolation}
\label{subsec: query_interpolation}

Our diffusion is applied on the reference points $\mathbf{r}_i$. As a first approach, we consider a DeformableDETR \cite{zhu2020deformable} architecture with $N$ learnable object query vectors which are assigned to their references by index. Thus, object query $\mathbf{q}_i$ may be placed in different $(x, y)$ locations depending on the sampling. While this approach works fairly well in practice, it clearly prohibits the model from learning positional information for query $\mathbf{q}_i$ simply because its position keeps changing during each training iteration.

\begin{figure}[tbp]
    \centering
    \includegraphics[width=\linewidth]{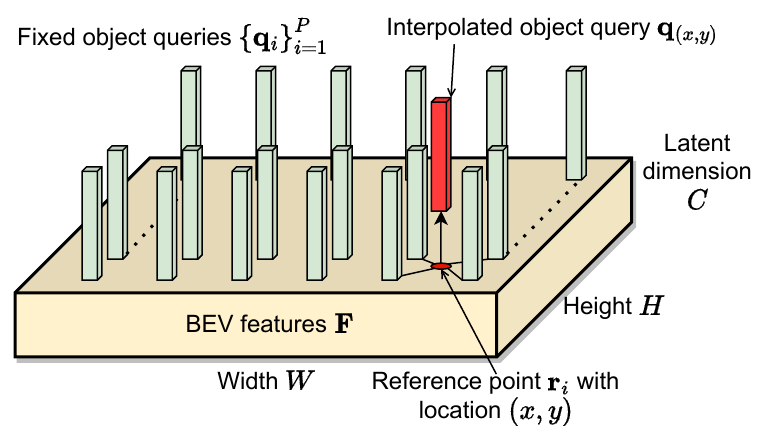}
    \captionsetup{aboveskip=0cm, belowskip=-0.4cm}
    \caption{\textbf{Query interpolation}. Our method learns a number of regularly-placed object queries, shown in green, which are fixed at test time. This allows the content of each object query to depend on its position in BEV. To accommodate the stochasticity required by the diffusion process we interpolate the object queries at the noisy locations, shown in red. This ensures that if we sample the same reference point many times, we will always obtain the same object query at that location.}
    \label{fig: query_interpolation}
\end{figure}

Instead, we propose to \emph{interpolate} the learned queries at the random sampled reference locations, as shown in Figure \ref{fig: query_interpolation}. We learn a grid of regularly-placed object queries, which we bilinearly interpolate at the reference points. This ensures that sampling the same location $\mathbf{r} = (x, y)$ will always yield the same object query $\mathbf{q}_{(x, y)}$. Additionally, this decouples the number of object queries at training time and at test time, since at training time one needs to learn $N$ queries, but at test time they can be interpolated at $N_{\text{test}}$ different locations.

In principle, one can also directly interpolate the BEV at the sampled locations, avoiding the use of learned object queries altogether. Our preliminary experiments showed that learning becomes prohibitively difficult in this case, owing to the diversity and nature of the BEV features. It is much easier if the model looks up the BEV features using cross-attention instead of starting from the BEV features as object queries.

\subsection{Additional method components}

\paragraph{Loss function.} The stochastic nature of the algorithm makes training very slow and difficult if we match predictions and ground truths in a one-to-one manner. To alleviate this, we employ many-to-one matching where many predictions are matched to each GT box. This speeds up training tremendously at the cost of having to post-process the predictions using non-maximum suppression (NMS).

Our loss function is given by
$$
\mathcal{L} = \lambda_{cls} \mathcal{L}_{cls} + \lambda_{reg} \mathcal{L}_{reg},
$$
where $\mathcal{L}_{cls}$ is the focal loss between predicted and target class probabilities \cite{lin2017focal} and $\mathcal{L}_{reg}$ is the $\ell_1$ loss between the predicted and GT box parameters. We do not employ a generalized IoU loss \cite{rezatofighi2019generalized}. The matching cost is the same as the loss function. For detection, the box parameters include the box center and dimensions, orientation, and velocity in the bird-eye-view plane:
$$
\mathbf{b} = (c_x, c_y, c_z, w, h, l, \sin \theta, \cos \theta, v_x, v_y).
$$

\begin{table*}[t!]
    \centering
    \begin{tabular}[width=\textwidth]{p{6.1cm} || c || c || c c c c c} \toprule[1.5pt]
         \textbf{Setting}&  mAP $\uparrow$& NDS $\uparrow$ & mATE $\downarrow$ & mASE $\downarrow$ & mAOE $\downarrow$ & mAVE $\downarrow$ & mAAE $\downarrow$ \\ \midrule[1pt]
         1. DiffusionDet \cite{chen2023diffusiondet} - ROI arch., box tokens& 0.3846& 0.4580 & 0.7420 & 0.3382 & 0.4475 & 0.5935 & 0.2222\\ 
         2. 1 + positional encodings &   0.3852& 0.4717 & 0.7357 & 0.2777 & 0.5040 & 0.5008 & 0.1911 \\ 
         3. DETR arch., random references&  0.3929& 0.4975 & 0.7172 & 0.2707 & 0.3835 & 0.4224  & 0.1963 \\ 
         4. 3 + \emph{simple} $n$-to-$1$ matching \& NMS &  0.4001 & 0.5203 & 0.6913 & 0.2710 & 0.3415 & 0.3540 & 0.1938 \\ 
         5. 3 + simOTA matching \& NMS & 0.3817 & 0.5138 & 0.6444 & \textbf{0.2641} & \textbf{0.3208} & \textbf{0.3422} & 0.1989 \\
         6. 4 + radial suppression & 0.4082 & 0.5203 & 0.6456 & 0.2704 & 0.3528 & 0.3739 & 0.1956 \\
         7. 4 + training with added fixed queries  & 0.4077 & 0.5215 & 0.6453 & 0.2696 & 0.3405 & 0.3747 & 0.1935 \\
         \rowcolor[gray]{0.9}
         8. 7 + radial suppression & \textbf{0.4088} & \textbf{0.5222} & \textbf{0.6437} & 0.2696 & 0.3395 & 0.3768 & 0.1922 \\
         \hline
         FCOS3D \cite{wang2021fcos3d} & 0.3430 & 0.4150 &  0.7250 & 0.2630 & 0.4220 & 1.292 & 0.153 \\
         PGD \cite{wang2022probabilistic} & 0.3690 & 0.4280 & 0.6830 & 0.2600 & 0.4390 & 1.2680 & 0.1850 \\
         DETR3D \cite{wang2022detr3d} & 0.3460 & 0.4250 & 0.7730 & 0.2680 & 0.3830 & 0.8420 & 0.2160 \\
         BEVFormer \cite{li2022bevformer}, permuted queries & 0.3976 & 0.5073 & 0.6809 & 0.2744 & 0.3722 & 0.3908 &  0.1962 \\
         BEVFormer, random reference points & 0.2997 & 0.4474 & 0.735 & 0.2765 & 0.3974 & 0.4179 & 0.1975  \\
         BEVFormer, deterministic &  0.4154& 0.5168 & 0.6715 & 0.2738 & 0.3691 & 0.3967 & 0.1981 \\ 
         \rowcolor[gray]{0.9}
         BEVFormer-Enh (ours) & \textbf{0.4189} & \textbf{0.5298} & \textbf{0.6319} & \textbf{0.2684} & \textbf{0.3283} & \textbf{0.3737} & \textbf{0.1945} \\
         \bottomrule[1.5pt]
    \end{tabular}
    \caption{\textbf{Model progression and results on the NuScenes \texttt{val} set}. We showcase how the model components and different architectures affect performance. Models numbered 1-8 are all evaluated with 3 DDIM steps and 900 queries.}
    \label{table: ablation_boxes_vs_object_queries}
\end{table*}

\textbf{Particle nature.} The many-to-one matching is crucial for our approach because it allows the model to learn gradient fields, or \emph{basins of attraction} around each GT box. This aspect, combined with the random reference points, allows us to look at this architecture as a particle DETR model where multiple particles, the references $\mathbf{r}_1, ..., \mathbf{r}_N$, can move freely and are attracted around the GT boxes. Through the self-attention layers, they can communicate similar to how the best location is globally shared in a particle swarm optimization \cite{kennedy1995particle}. The DDIM denoising \cite{song2020denoising} steps then provide opportunities to refine, renew, or prune the particles, based on their confidence. Additionally, the number of particles which end up on top of a target object can provide a rudimentary measure about the uncertainty of our perceptions at that BEV location. We cannot refer to the search tokens of DETR models \cite{carion2020end, zhu2020deformable} as dynamic because they are fixed and do not allow for sequential refinement.
\section{Experiments}
\label{sec: experiments}

\paragraph{NuScenes dataset.} We evaluate our approach on the large-scale NuScenes dataset \cite{caesar2020nuscenes}, comprising almost 1.4 million annotated 3D bounding boxes, across 1000 scenes. There are 23 semantic classes of which 10 are evaluated. The frequency of the images is 2 Hz. The main metrics of interest are the Mean Average Precision (mAP) and, more importantly, the NuScenes Detection Score (NDS). 

For the mAP detections are calculated by greedily assigning predictions to targets only based on the distance between the predicted and GT centers. There are four distance thresholds - 0.5, 1, 2, and 4 meters. The mAP is calculated as the average precision over 100 recall percentiles and is further averaged over all 10 detectable classes and over these 4 distance thresholds.

At evaluation time, once the assignment between predicted boxes and targets is completed, one can calculate various true positive metrics - translation error (mATE), scale error (mASE), box orientation (mAOE), velocity (mAVE), attribute error (mAAE) - over the matched pairs. These are weighted together with the mAP to form the NDS metric. It has been claimed that the NDS metric is more realistic in terms of real-life driving performance than the mAP \cite{schreier2023offline}.

\subsection{Comparison with baselines}
\label{subsection: comparison_with_baselines}

In our experiments we compare against the following state-of-the-art models:

\begin{enumerate}
    \item DiffusionDet \cite{chen2023diffusiondet}, which we modify minimally and utilize directly in BEV as our main baseline,
    \item DeformableDETR \cite{zhu2020deformable}, as used in BEVFormer \cite{li2022bevformer}, a state-of-the-art deterministic detector which already greatly outperforms the vanilla DETR model \cite{carion2020end}.
\end{enumerate}

\textbf{Baseline.} Table \ref{table: ablation_boxes_vs_object_queries} shows our main results. We rely on BEVFormer's encoder to project the images into the top-down view. Since the original DiffusionDet works only on axis-aligned boxes, we modify it by adopting rotated ROI pooling similar to \cite{xie2021oriented}. The architecture follows a six stage RCNN \cite{girshick2014rich} decoder where each stage takes the BEV features and a number of rotated boxes in BEV, parameterized as $(c_x, c_y, w, h, \theta)$. The BEV features falling into the rotated box are aggregated and deformable convolutions \cite{dai2017deformable} are applied to model instance interactions between different boxes. Each stage outputs corrections which are applied to the current boxes to produce the subsequent-stage boxes. Overall, applying DiffusionDet directly in BEV yields good performance compared to reference models \cite{wang2021fcos3d, wang2022probabilistic, wang2022detr3d} but inferior compared to the deterministic BEVFormer.

\textbf{Positional encodings.} It is common to encounter certain classes more often in some positions relative to the ego-vehicle, e.g. pedestrians appear in front of the car less often than at the side of the car. The ROI-based architecture does not consider the absolute locations of the boxes in BEV, which motivates us to use sinusoidal positional encodings \cite{vaswani2017attention}, which we concatenate to the aggregated BEV features for each box token. This improves performance but is still insufficient compared to BEVFormer.

\textbf{Global features to address sparsity.} ROI-based architectures emphasize the local features inside each box. Such a prior may be sufficient on datasets like COCO \cite{lin2014microsoft}, but for smaller objects we argue that more global features are needed. This motivates us to consider a DETR-based architecture where instead of boxes and ROI-pooling we have object queries and attention. Now, each stage first applies self-attention over the object queries, thereby considering their relative position and content, and then applies cross-attention over the BEV. This cross-attention has potentially unlimited view and can aggregate more global BEV features for each token than the ROI architecture.

\textbf{Many-to-one matching}. At this point, even though global features and positions are considered, we found that with random reference points $\mathbf{r}_1, ..., \mathbf{r}_N$, the supervisory signal when matching in a one-to-one fashion is simply too weak. Thus, we experiment with two many-to-one matching strategies. The first we call \emph{simple} $N$-to-1 because it simply repeats the GT boxes a number of times, stacking them on top of each other, and then applies the linear sum assignment solver for matching. For the second strategy we use the SimOTA \cite{ge2021yolox} approximation to optimal transport assignment \cite{ge2021ota}, which matches a variable number of predictions to each target. 

\textbf{Detection accuracy.} The results show that our diffusion-based Particle-DETR achieves good performance and noticeably outperforms the baseline DiffusionDet \cite{chen2023diffusion} on both mAP and NDS. Even more, its performance is comparable to that of deterministic approaches like BEVFormer \cite{li2022bevformer}. Our generative approach achieves higher NDS, showing that once a detection has been established, the predicted box dimensions, orientation, and velocity are, on average, more accurate. Yet, a small gap of about 1.6 mAP points remains.

\textbf{Enhancement with static references}. The random references allow the model to learn basins of attraction around each GT center. However, nothing prevents us from utilizing fixed references as well, which yield higher precision. Thus, we further experiment with a setup in which we train with two sets of references - one random, coming from the diffusion process, and one static. In turn, the two reference sets result in two sets of queries - one where the queries are interpolated at the random locations (cf. Subsection \ref{subsec: query_interpolation}), and one where the queries are learned and fixed, as in \cite{li2022bevformer, zhu2020deformable}. Since the decoder is shared, the joint training captures any synergies between the random and fixed queries, improving the performance of both. At test time, to keep the number of queries comparable to previous models, we can use either only the diffusion queries or the static ones. Using the diffusion queries we obtain our final Particle-DETR model. Using the static ones we obtain an enhanced BEVFormer which we call BEVFormer-Enh.

\subsection{Implementation details}
The implementation of our Particle-DETR is straightforward and follows that of BEVFormer \cite{li2022bevformer}. We train the model for the same number of iterations as BEVFormer and the number of parameters is similar. The training hyperparameters are shown in Table \ref{table: hyperparameters} and pseudocodes for the train-test behaviour can be found in Algorithms \ref{algo: pseudocode_train} and \ref{algo: pseudocode_test}.

\textbf{Gradient detachment.} To further facilitate training, we equip each decoder layer with \emph{look forward twice} updates \cite{zhang2022dino}, where the reference points for each decoder layer are not detached from the computation graph when computing the next-layer reference points during the forward pass.

\textbf{Filtering of predictions.} At training time, the many-to-one matching helps to learn the basins of attraction around each GT center. However, at test time it results in many false positives, as can be seen from Section \ref{supp: additional_experiments} in the supplementary materials. Thus, we employ NMS and also utilize a small score threshold which filters any predictions with confidence below it.

\begin{figure}[bp]
    \centering
    \includegraphics[width=\linewidth]{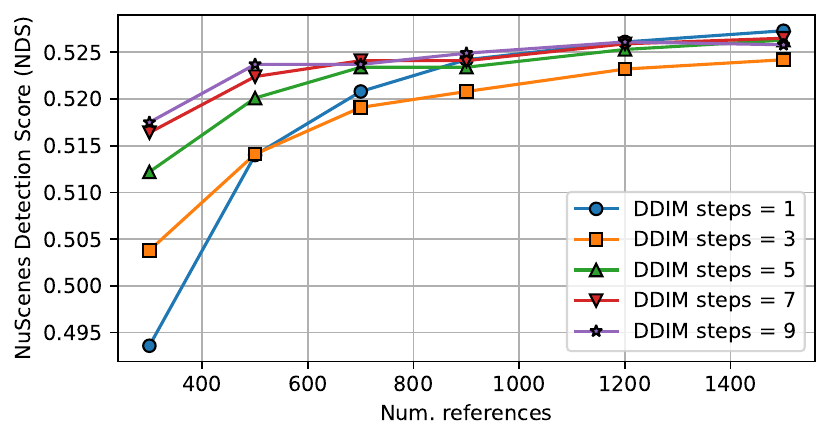}
    \captionsetup{aboveskip=0cm, belowskip=-0.1cm}
    \caption{\textbf{Effect of number of references on NDS}. Holding the number of DDIM steps fixed, NDS increases as the number of random references increases.}
    \label{fig: num_refs_vs_nds}
\end{figure}

\textbf{Radial suppression.} We found that very small objects like traffic cones do not overlap and are missed by NMS. For that reason, we introduced \emph{radial suppression} to further filter out the boxes. In essence, we first order the predictions by decreasing confidence. Then for the most confident ones, we sequentially replace them with weighted averages of their close-by boxes which, in turn, are filtered:
$$
    \mathbf{b}_i = \frac{\sum_{k} \mathbf{b}_k \pi_k}{\sum_{k} \pi_k}, \  \\
    \forall k: \sqrt{(c_{x, i} - c_{x, k})^2 + (c_{y, i} - c_{y, k})^2 } < r.
$$

Here $c_{x, k}$ is the $x$-coordinate of the center of the $k$-th box, and $\pi_k$ is the confidence for that box. We implement radial suppression independently for each semantic class.

\subsection{Additional properties}

\begin{figure}[bp]
    \centering
    \includegraphics[width=\linewidth]{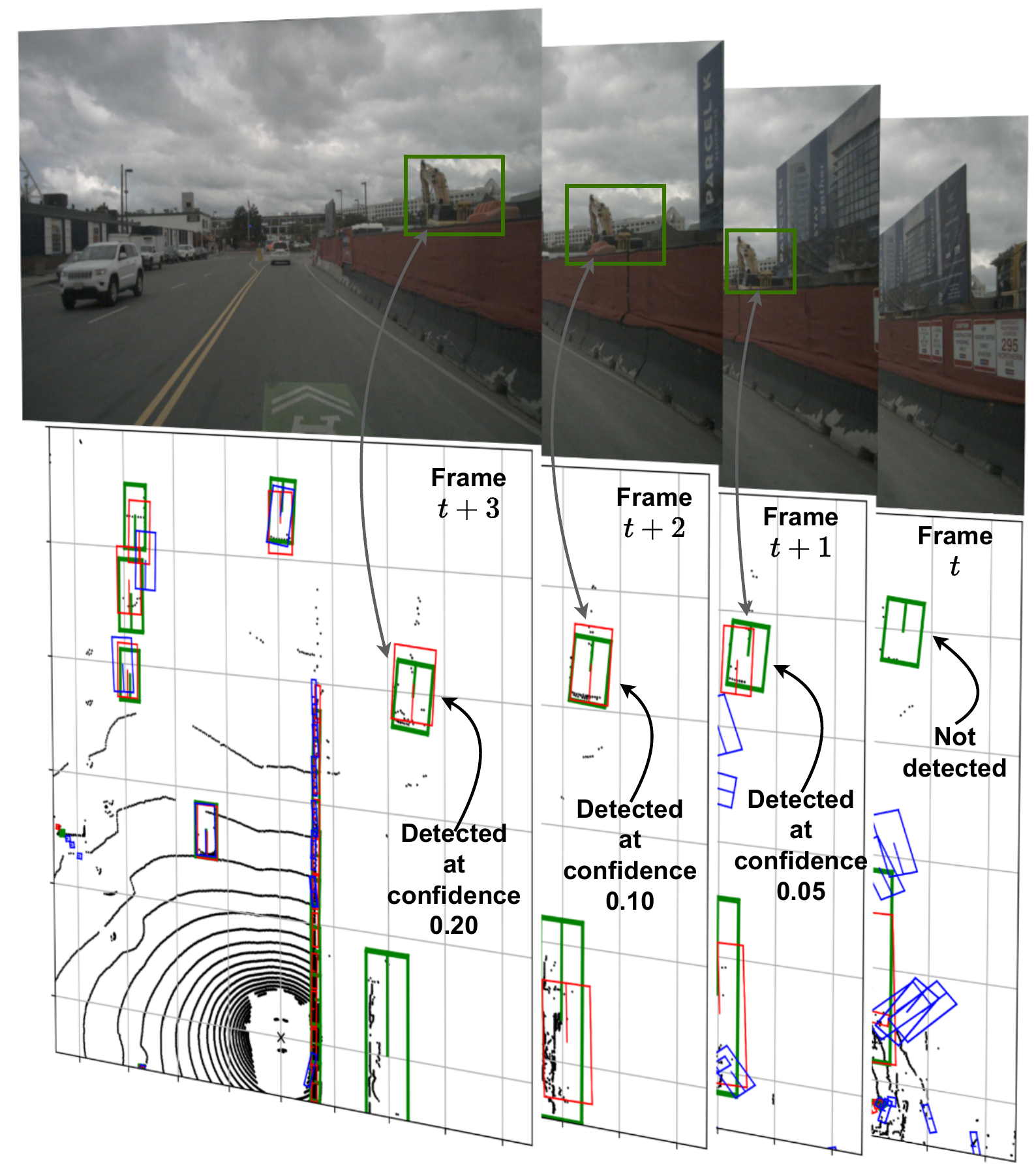}
    \caption{\textbf{Sample predictions in BEV}. Green boxes are ground-truths, red are predicted by our Particle-DETR, and blue is predicted by BEVFormer, compared to which we detect earlier in the frames and more confidently, even for less-common objects.}
    \label{fig: sample_prediction_in_bev}
\end{figure}

\textbf{Flexibility.} The architecture of our Particle-DETR allows us to train it with one number of queries but evaluate with a different number. Additionally, the number of DDIM steps \cite{song2020denoising} allows us to further trade-off accuracy and compute. Figure \ref{fig: num_refs_vs_nds} shows that both increasing the number of DDIM steps and the number of particles used improves performance. With 900 references it only takes a single DDIM step to outperform BEVFormer on NDS.

\textbf{Stochastic nature of results.} Since we rely on randomly sampled initial reference points, the outputs of our method are stochastic. Table \ref{table: average_metrics} shows statistics over 10 test runs. We note that performance is very consistent across the runs.

\subsection{Qualitative study}
Here we perform a qualitative comparison between our predictions and those of BEVFormer \cite{li2022bevformer}. In general, the higher NDS which results from the diffusion process  makes our detections more precise in terms of location, size, and orientation, which can be particularly beneficial for very small objects (e.g. traffic cones) near the car. On some scenes our method recognizes even partially-occluded objects earlier and more confidently, as shown in Figure \ref{fig: sample_prediction_in_bev}.

It is common for models to struggle with accurate estimation of the dimensions of large objects like buses and trucks. This is because they obscure the camera's field of view considerably, making it hard to estimate where the object ends. We notice that in some scenes our method improves noticeably in this regard. Further visualizations and analysis can be found in the supplementary materials.

\begin{table}[tbp]
    \centering
    \begin{tabular}{ l | c c c } \toprule[1.5pt]
         \textbf{Metric} &  BEVFormer & Ours (stoc.) & Ours \\
         \midrule[1pt]
         NDS $\uparrow$ & 0.5168 & 0.5271 (0.0002) & \textbf{0.5287}  \\
         mAP $\uparrow$  & 0.4154 & 0.4163 (0.0003) & \textbf{0.4184}  \\ 
         mATE $\downarrow$ & 0.6715 & 0.6415 (0.0008) & \textbf{0.6386} \\
         mASE $\downarrow$ & 0.2738 & 0.2689 (0.0002) & \textbf{0.2686} \\
         mAOE $\downarrow$ & 0.3691 & 0.3390 (0.0009) & \textbf{0.3362} \\
         mAVE $\downarrow$ & 0.4179 & \textbf{0.3688} (0.0010) & \textbf{0.3688} \\
        mAAE $\downarrow$ & 0.1981 & \textbf{0.1920} (0.0006) & 0.1931 \\ 
        \bottomrule[1.5pt]
    \end{tabular}
    \captionsetup{aboveskip=0.1cm, belowskip=-0.4cm}
    \caption{\textbf{Performance statistics on the NuScenes \texttt{val} set}. We compare our stochastic Particle-DETR (col. 3), evaluated with 1500 queries and 1 DDIM step, and the deterministic BEVFormer-Enh (col. 4) to the original BEVFormer. The standard deviations for the random methods are shown in parentheses.}
    \label{table: average_metrics}
\end{table}
\section{Discussion}
\label{sec: discussion}

\textbf{Precision in generative models.} Learning a distribution for the bounding boxes given the BEV features stands in stark contrast to text-to-image tasks. One should recognize that even for a very detailed text prompt, there are many corresponding valid images. Perturbing the pixel values will not change the description significantly. Thus, text-to-image tasks tolerate a large amount of variation in the generated samples. Object detection, however, requires precision in the outputs. Hence, adjusting for number of queries, we find a small performance gap in mAP with respect to deterministic approaches natural, as the random reference inputs will always induce a distribution on the outputs.

\textbf{Uncertainty.} One benefit of learning a distribution over the boxes is that this provides a rudimentary way to understand their uncertainty. Unfortunately, it is likely that it mixes epistemic uncertainty resulting from the estimated model parameters and aleatoric uncertainty related to the randomness of the boxes themselves. Heatmaps showing estimated box distributions are available in the supplementary materials, along with further discussion.

\section{Conclusion}
\label{sec: conclusion}

In this work we explored diffusion-based generative models for 3D object detection in BEV. We have shown that naively using previous approaches yields a performance gap. To close it, we adopt a transformer-based architecture and a specific query interpolation module to facilitate the model in learning positional information even in the presence of diffusion. We formulate the diffusion process as diffusion over particles, which yields a unique interpretation based on particle methods. Our approach greatly improves on previous generative methods and achieves comparable results to strong deterministic ones.
\newpage

\section{Aknowledgements}
This research was partially funded by the Ministry of Education and
Science of Bulgaria (support for INSAIT, part of the Bulgarian National
Roadmap for Research Infrastructure).

{
    \small
    \bibliographystyle{ieeenat_fullname}
    \bibliography{main}

\begin{thebibliography}{72}
\providecommand{\natexlab}[1]{#1}
\providecommand{\url}[1]{\texttt{#1}}
\expandafter\ifx\csname urlstyle\endcsname\relax
  \providecommand{\doi}[1]{doi: #1}\else
  \providecommand{\doi}{doi: \begingroup \urlstyle{rm}\Url}\fi

\bibitem[Barquero et~al.(2023)Barquero, Escalera, and Palmero]{barquero2023belfusion}
German Barquero, Sergio Escalera, and Cristina Palmero.
\newblock Belfusion: Latent diffusion for behavior-driven human motion prediction.
\newblock In \emph{Proceedings of the IEEE/CVF International Conference on Computer Vision}, pages 2317--2327, 2023.

\bibitem[Brazil and Liu(2019)]{brazil2019m3d}
Garrick Brazil and Xiaoming Liu.
\newblock M3d-rpn: Monocular 3d region proposal network for object detection.
\newblock In \emph{Proceedings of the IEEE/CVF International Conference on Computer Vision}, pages 9287--9296, 2019.

\bibitem[Brazil et~al.(2020)Brazil, Pons-Moll, Liu, and Schiele]{brazil2020kinematic}
Garrick Brazil, Gerard Pons-Moll, Xiaoming Liu, and Bernt Schiele.
\newblock Kinematic 3d object detection in monocular video.
\newblock In \emph{Computer Vision--ECCV 2020: 16th European Conference, Glasgow, UK, August 23--28, 2020, Proceedings, Part XXIII 16}, pages 135--152. Springer, 2020.

\bibitem[Caesar et~al.(2020)Caesar, Bankiti, Lang, Vora, Liong, Xu, Krishnan, Pan, Baldan, and Beijbom]{caesar2020nuscenes}
Holger Caesar, Varun Bankiti, Alex~H Lang, Sourabh Vora, Venice~Erin Liong, Qiang Xu, Anush Krishnan, Yu Pan, Giancarlo Baldan, and Oscar Beijbom.
\newblock nuscenes: A multimodal dataset for autonomous driving.
\newblock In \emph{Proceedings of the IEEE/CVF conference on computer vision and pattern recognition}, pages 11621--11631, 2020.

\bibitem[Carion et~al.(2020)Carion, Massa, Synnaeve, Usunier, Kirillov, and Zagoruyko]{carion2020end}
Nicolas Carion, Francisco Massa, Gabriel Synnaeve, Nicolas Usunier, Alexander Kirillov, and Sergey Zagoruyko.
\newblock End-to-end object detection with transformers.
\newblock In \emph{European conference on computer vision}, pages 213--229. Springer, 2020.

\bibitem[Chai et~al.(2019)Chai, Sapp, Bansal, and Anguelov]{chai2019multipath}
Yuning Chai, Benjamin Sapp, Mayank Bansal, and Dragomir Anguelov.
\newblock Multipath: Multiple probabilistic anchor trajectory hypotheses for behavior prediction.
\newblock \emph{arXiv preprint arXiv:1910.05449}, 2019.

\bibitem[Chen et~al.(2023{\natexlab{a}})Chen, Chen, Wang, Zhang, Yao, Feng, Han, Ding, Zeng, and Wang]{chen2023group}
Qiang Chen, Xiaokang Chen, Jian Wang, Shan Zhang, Kun Yao, Haocheng Feng, Junyu Han, Errui Ding, Gang Zeng, and Jingdong Wang.
\newblock Group detr: Fast detr training with group-wise one-to-many assignment.
\newblock In \emph{Proceedings of the IEEE/CVF International Conference on Computer Vision}, pages 6633--6642, 2023{\natexlab{a}}.

\bibitem[Chen et~al.(2023{\natexlab{b}})Chen, Sun, Song, and Luo]{chen2023diffusiondet}
Shoufa Chen, Peize Sun, Yibing Song, and Ping Luo.
\newblock Diffusiondet: Diffusion model for object detection.
\newblock In \emph{Proceedings of the IEEE/CVF International Conference on Computer Vision}, pages 19830--19843, 2023{\natexlab{b}}.

\bibitem[Chen et~al.(2023{\natexlab{c}})Chen, Gao, Xiang, and Lin]{chen2023diffusion}
Zhennan Chen, Rongrong Gao, Tian-Zhu Xiang, and Fan Lin.
\newblock Diffusion model for camouflaged object detection.
\newblock \emph{arXiv preprint arXiv:2308.00303}, 2023{\natexlab{c}}.

\bibitem[Chung et~al.(2022)Chung, Sim, and Ye]{chung2022come}
Hyungjin Chung, Byeongsu Sim, and Jong~Chul Ye.
\newblock Come-closer-diffuse-faster: Accelerating conditional diffusion models for inverse problems through stochastic contraction.
\newblock In \emph{Proceedings of the IEEE/CVF Conference on Computer Vision and Pattern Recognition}, pages 12413--12422, 2022.

\bibitem[Croitoru et~al.(2023)Croitoru, Hondru, Ionescu, and Shah]{croitoru2023diffusion}
Florinel-Alin Croitoru, Vlad Hondru, Radu~Tudor Ionescu, and Mubarak Shah.
\newblock Diffusion models in vision: A survey.
\newblock \emph{IEEE Transactions on Pattern Analysis and Machine Intelligence}, 2023.

\bibitem[Cui et~al.(2023)Cui, Casas, Wong, Suo, and Urtasun]{cui2023gorela}
Alexander Cui, Sergio Casas, Kelvin Wong, Simon Suo, and Raquel Urtasun.
\newblock Gorela: Go relative for viewpoint-invariant motion forecasting.
\newblock In \emph{2023 IEEE International Conference on Robotics and Automation (ICRA)}, pages 7801--7807. IEEE, 2023.

\bibitem[Dai et~al.(2017)Dai, Qi, Xiong, Li, Zhang, Hu, and Wei]{dai2017deformable}
Jifeng Dai, Haozhi Qi, Yuwen Xiong, Yi Li, Guodong Zhang, Han Hu, and Yichen Wei.
\newblock Deformable convolutional networks.
\newblock In \emph{Proceedings of the IEEE international conference on computer vision}, pages 764--773, 2017.

\bibitem[Dai et~al.(2021)Dai, Chen, Yang, Zhang, Yuan, and Zhang]{dai2021dynamic}
Xiyang Dai, Yinpeng Chen, Jianwei Yang, Pengchuan Zhang, Lu Yuan, and Lei Zhang.
\newblock Dynamic detr: End-to-end object detection with dynamic attention.
\newblock In \emph{Proceedings of the IEEE/CVF International Conference on Computer Vision}, pages 2988--2997, 2021.

\bibitem[Dendorfer et~al.(2022)Dendorfer, Yugay, Osep, and Leal-Taix{\'e}]{dendorfer2022quo}
Patrick Dendorfer, Vladimir Yugay, Aljosa Osep, and Laura Leal-Taix{\'e}.
\newblock Quo vadis: Is trajectory forecasting the key towards long-term multi-object tracking?
\newblock \emph{Advances in Neural Information Processing Systems}, 35:\penalty0 15657--15671, 2022.

\bibitem[Deng et~al.(2009)Deng, Dong, Socher, Li, Li, and Fei-Fei]{deng2009imagenet}
Jia Deng, Wei Dong, Richard Socher, Li-Jia Li, Kai Li, and Li Fei-Fei.
\newblock Imagenet: A large-scale hierarchical image database.
\newblock In \emph{2009 IEEE conference on computer vision and pattern recognition}, pages 248--255. Ieee, 2009.

\bibitem[Everingham et~al.(2010)Everingham, Van~Gool, Williams, Winn, and Zisserman]{everingham2010pascal}
Mark Everingham, Luc Van~Gool, Christopher~KI Williams, John Winn, and Andrew Zisserman.
\newblock The pascal visual object classes (voc) challenge.
\newblock \emph{International journal of computer vision}, 88:\penalty0 303--338, 2010.

\bibitem[Feng et~al.(2023)Feng, Gao, Tse, Ma, and Chang]{feng2023diffpose}
Runyang Feng, Yixing Gao, Tze Ho~Elden Tse, Xueqing Ma, and Hyung~Jin Chang.
\newblock Diffpose: Spatiotemporal diffusion model for video-based human pose estimation.
\newblock In \emph{Proceedings of the IEEE/CVF International Conference on Computer Vision}, pages 14861--14872, 2023.

\bibitem[Ge et~al.(2021{\natexlab{a}})Ge, Liu, Li, Yoshie, and Sun]{ge2021ota}
Zheng Ge, Songtao Liu, Zeming Li, Osamu Yoshie, and Jian Sun.
\newblock Ota: Optimal transport assignment for object detection.
\newblock In \emph{Proceedings of the IEEE/CVF Conference on Computer Vision and Pattern Recognition}, pages 303--312, 2021{\natexlab{a}}.

\bibitem[Ge et~al.(2021{\natexlab{b}})Ge, Liu, Wang, Li, and Sun]{ge2021yolox}
Zheng Ge, Songtao Liu, Feng Wang, Zeming Li, and Jian Sun.
\newblock Yolox: Exceeding yolo series in 2021.
\newblock \emph{arXiv preprint arXiv:2107.08430}, 2021{\natexlab{b}}.

\bibitem[Girshick et~al.(2014)Girshick, Donahue, Darrell, and Malik]{girshick2014rich}
Ross Girshick, Jeff Donahue, Trevor Darrell, and Jitendra Malik.
\newblock Rich feature hierarchies for accurate object detection and semantic segmentation.
\newblock In \emph{Proceedings of the IEEE conference on computer vision and pattern recognition}, pages 580--587, 2014.

\bibitem[Gwak et~al.(2022)Gwak, Savarese, and Bohg]{gwak2022minkowski}
JunYoung Gwak, Silvio Savarese, and Jeannette Bohg.
\newblock Minkowski tracker: A sparse spatio-temporal r-cnn for joint object detection and tracking.
\newblock \emph{arXiv preprint arXiv:2208.10056}, 2022.

\bibitem[He et~al.(2016)He, Zhang, Ren, and Sun]{he2016deep}
Kaiming He, Xiangyu Zhang, Shaoqing Ren, and Jian Sun.
\newblock Deep residual learning for image recognition.
\newblock In \emph{Proceedings of the IEEE conference on computer vision and pattern recognition}, pages 770--778, 2016.

\bibitem[He et~al.(2017)He, Gkioxari, Doll{\'a}r, and Girshick]{he2017mask}
Kaiming He, Georgia Gkioxari, Piotr Doll{\'a}r, and Ross Girshick.
\newblock Mask r-cnn.
\newblock In \emph{Proceedings of the IEEE international conference on computer vision}, pages 2961--2969, 2017.

\bibitem[Ho et~al.(2020)Ho, Jain, and Abbeel]{ho2020denoising}
Jonathan Ho, Ajay Jain, and Pieter Abbeel.
\newblock Denoising diffusion probabilistic models.
\newblock \emph{Advances in neural information processing systems}, 33:\penalty0 6840--6851, 2020.

\bibitem[Hu et~al.(2021)Hu, Murez, Mohan, Dudas, Hawke, Badrinarayanan, Cipolla, and Kendall]{hu2021fiery}
Anthony Hu, Zak Murez, Nikhil Mohan, Sof{\'\i}a Dudas, Jeffrey Hawke, Vijay Badrinarayanan, Roberto Cipolla, and Alex Kendall.
\newblock Fiery: Future instance prediction in bird's-eye view from surround monocular cameras.
\newblock In \emph{Proceedings of the IEEE/CVF International Conference on Computer Vision}, pages 15273--15282, 2021.

\bibitem[Hu et~al.(2022)Hu, Chen, Wu, Li, Yan, and Tao]{hu2022st}
Shengchao Hu, Li Chen, Penghao Wu, Hongyang Li, Junchi Yan, and Dacheng Tao.
\newblock St-p3: End-to-end vision-based autonomous driving via spatial-temporal feature learning.
\newblock In \emph{European Conference on Computer Vision}, pages 533--549. Springer, 2022.

\bibitem[Huang and Huang(2022)]{huang2022bevdet4d}
Junjie Huang and Guan Huang.
\newblock Bevdet4d: Exploit temporal cues in multi-camera 3d object detection.
\newblock \emph{arXiv preprint arXiv:2203.17054}, 2022.

\bibitem[Jiang et~al.(2023)Jiang, Zhang, Miao, Zhu, Gao, Hu, and Jiang]{jiang2023polarformer}
Yanqin Jiang, Li Zhang, Zhenwei Miao, Xiatian Zhu, Jin Gao, Weiming Hu, and Yu-Gang Jiang.
\newblock Polarformer: Multi-camera 3d object detection with polar transformer.
\newblock In \emph{Proceedings of the AAAI Conference on Artificial Intelligence}, pages 1042--1050, 2023.

\bibitem[Jin et~al.(2023)Jin, Li, Cheng, Li, Ji, Liu, Yuan, and Chen]{jin2023diffusionret}
Peng Jin, Hao Li, Zesen Cheng, Kehan Li, Xiangyang Ji, Chang Liu, Li Yuan, and Jie Chen.
\newblock Diffusionret: Generative text-video retrieval with diffusion model.
\newblock \emph{arXiv preprint arXiv:2303.09867}, 2023.

\bibitem[Kennedy and Eberhart(1995)]{kennedy1995particle}
James Kennedy and Russell Eberhart.
\newblock Particle swarm optimization.
\newblock In \emph{Proceedings of ICNN'95-international conference on neural networks}, pages 1942--1948. IEEE, 1995.

\bibitem[Li et~al.(2022{\natexlab{a}})Li, Zhang, Liu, Guo, Ni, and Zhang]{li2022dn}
Feng Li, Hao Zhang, Shilong Liu, Jian Guo, Lionel~M Ni, and Lei Zhang.
\newblock Dn-detr: Accelerate detr training by introducing query denoising.
\newblock In \emph{Proceedings of the IEEE/CVF Conference on Computer Vision and Pattern Recognition}, pages 13619--13627, 2022{\natexlab{a}}.

\bibitem[Li et~al.(2022{\natexlab{b}})Li, Sima, Dai, Wang, Lu, Wang, Xie, Li, Deng, Tian, et~al.]{li2022delving}
Hongyang Li, Chonghao Sima, Jifeng Dai, Wenhai Wang, Lewei Lu, Huijie Wang, Enze Xie, Zhiqi Li, Hanming Deng, Hao Tian, et~al.
\newblock Delving into the devils of bird's-eye-view perception: A review, evaluation and recipe.
\newblock \emph{arXiv preprint arXiv:2209.05324}, 2022{\natexlab{b}}.

\bibitem[Li et~al.(2022{\natexlab{c}})Li, Wang, Li, Xie, Sima, Lu, Qiao, and Dai]{li2022bevformer}
Zhiqi Li, Wenhai Wang, Hongyang Li, Enze Xie, Chonghao Sima, Tong Lu, Yu Qiao, and Jifeng Dai.
\newblock Bevformer: Learning bird’s-eye-view representation from multi-camera images via spatiotemporal transformers.
\newblock In \emph{European conference on computer vision}, pages 1--18. Springer, 2022{\natexlab{c}}.

\bibitem[Lin et~al.(2014)Lin, Maire, Belongie, Hays, Perona, Ramanan, Doll{\'a}r, and Zitnick]{lin2014microsoft}
Tsung-Yi Lin, Michael Maire, Serge Belongie, James Hays, Pietro Perona, Deva Ramanan, Piotr Doll{\'a}r, and C~Lawrence Zitnick.
\newblock Microsoft coco: Common objects in context.
\newblock In \emph{Computer Vision--ECCV 2014: 13th European Conference, Zurich, Switzerland, September 6-12, 2014, Proceedings, Part V 13}, pages 740--755. Springer, 2014.

\bibitem[Lin et~al.(2017{\natexlab{a}})Lin, Doll{\'a}r, Girshick, He, Hariharan, and Belongie]{lin2017feature}
Tsung-Yi Lin, Piotr Doll{\'a}r, Ross Girshick, Kaiming He, Bharath Hariharan, and Serge Belongie.
\newblock Feature pyramid networks for object detection.
\newblock In \emph{Proceedings of the IEEE conference on computer vision and pattern recognition}, pages 2117--2125, 2017{\natexlab{a}}.

\bibitem[Lin et~al.(2017{\natexlab{b}})Lin, Goyal, Girshick, He, and Doll{\'a}r]{lin2017focal}
Tsung-Yi Lin, Priya Goyal, Ross Girshick, Kaiming He, and Piotr Doll{\'a}r.
\newblock Focal loss for dense object detection.
\newblock In \emph{Proceedings of the IEEE international conference on computer vision}, pages 2980--2988, 2017{\natexlab{b}}.

\bibitem[Liu et~al.(2023)Liu, Li, Dinh, Jiang, Shah, and Xu]{liu2023diffusion}
Daochang Liu, Qiyue Li, Anh-Dung Dinh, Tingting Jiang, Mubarak Shah, and Chang Xu.
\newblock Diffusion action segmentation.
\newblock In \emph{Proceedings of the IEEE/CVF International Conference on Computer Vision}, pages 10139--10149, 2023.

\bibitem[Liu et~al.(2022)Liu, Li, Zhang, Yang, Qi, Su, Zhu, and Zhang]{liu2022dab}
Shilong Liu, Feng Li, Hao Zhang, Xiao Yang, Xianbiao Qi, Hang Su, Jun Zhu, and Lei Zhang.
\newblock Dab-detr: Dynamic anchor boxes are better queries for detr.
\newblock \emph{arXiv preprint arXiv:2201.12329}, 2022.

\bibitem[Liu et~al.(2021)Liu, Lin, Cao, Hu, Wei, Zhang, Lin, and Guo]{liu2021swin}
Ze Liu, Yutong Lin, Yue Cao, Han Hu, Yixuan Wei, Zheng Zhang, Stephen Lin, and Baining Guo.
\newblock Swin transformer: Hierarchical vision transformer using shifted windows.
\newblock In \emph{Proceedings of the IEEE/CVF international conference on computer vision}, pages 10012--10022, 2021.

\bibitem[Luo et~al.(2018)Luo, Yang, and Urtasun]{luo2018fast}
Wenjie Luo, Bin Yang, and Raquel Urtasun.
\newblock Fast and furious: Real time end-to-end 3d detection, tracking and motion forecasting with a single convolutional net.
\newblock In \emph{Proceedings of the IEEE conference on Computer Vision and Pattern Recognition}, pages 3569--3577, 2018.

\bibitem[Mao et~al.(2023{\natexlab{a}})Mao, Shi, Wang, and Li]{mao20233d}
Jiageng Mao, Shaoshuai Shi, Xiaogang Wang, and Hongsheng Li.
\newblock 3d object detection for autonomous driving: A comprehensive survey.
\newblock \emph{International Journal of Computer Vision}, pages 1--55, 2023{\natexlab{a}}.

\bibitem[Mao et~al.(2023{\natexlab{b}})Mao, Xu, Zhu, Chen, and Wang]{mao2023leapfrog}
Weibo Mao, Chenxin Xu, Qi Zhu, Siheng Chen, and Yanfeng Wang.
\newblock Leapfrog diffusion model for stochastic trajectory prediction.
\newblock In \emph{Proceedings of the IEEE/CVF Conference on Computer Vision and Pattern Recognition}, pages 5517--5526, 2023{\natexlab{b}}.

\bibitem[Nichol and Dhariwal(2021)]{nichol2021improved}
Alexander~Quinn Nichol and Prafulla Dhariwal.
\newblock Improved denoising diffusion probabilistic models.
\newblock In \emph{International Conference on Machine Learning}, pages 8162--8171. PMLR, 2021.

\bibitem[Peng et~al.(2023{\natexlab{a}})Peng, Hu, Ke, and Liu]{peng2023diffusion}
Duo Peng, Ping Hu, Qiuhong Ke, and Jun Liu.
\newblock Diffusion-based image translation with label guidance for domain adaptive semantic segmentation.
\newblock In \emph{Proceedings of the IEEE/CVF International Conference on Computer Vision}, pages 808--820, 2023{\natexlab{a}}.

\bibitem[Peng et~al.(2023{\natexlab{b}})Peng, Chen, Fu, Liang, and Cheng]{peng2023bevsegformer}
Lang Peng, Zhirong Chen, Zhangjie Fu, Pengpeng Liang, and Erkang Cheng.
\newblock Bevsegformer: Bird's eye view semantic segmentation from arbitrary camera rigs.
\newblock In \emph{Proceedings of the IEEE/CVF Winter Conference on Applications of Computer Vision}, pages 5935--5943, 2023{\natexlab{b}}.

\bibitem[Philion and Fidler(2020)]{philion2020lift}
Jonah Philion and Sanja Fidler.
\newblock Lift, splat, shoot: Encoding images from arbitrary camera rigs by implicitly unprojecting to 3d.
\newblock In \emph{Computer Vision--ECCV 2020: 16th European Conference, Glasgow, UK, August 23--28, 2020, Proceedings, Part XIV 16}, pages 194--210. Springer, 2020.

\bibitem[Qin et~al.(2023)Qin, Chen, Chen, Chen, and Li]{qin2023unifusion}
Zequn Qin, Jingyu Chen, Chao Chen, Xiaozhi Chen, and Xi Li.
\newblock Unifusion: Unified multi-view fusion transformer for spatial-temporal representation in bird's-eye-view.
\newblock In \emph{Proceedings of the IEEE/CVF International Conference on Computer Vision}, pages 8690--8699, 2023.

\bibitem[Reading et~al.(2021)Reading, Harakeh, Chae, and Waslander]{reading2021categorical}
Cody Reading, Ali Harakeh, Julia Chae, and Steven~L Waslander.
\newblock Categorical depth distribution network for monocular 3d object detection.
\newblock In \emph{Proceedings of the IEEE/CVF Conference on Computer Vision and Pattern Recognition}, pages 8555--8564, 2021.

\bibitem[Rezatofighi et~al.(2019)Rezatofighi, Tsoi, Gwak, Sadeghian, Reid, and Savarese]{rezatofighi2019generalized}
Hamid Rezatofighi, Nathan Tsoi, JunYoung Gwak, Amir Sadeghian, Ian Reid, and Silvio Savarese.
\newblock Generalized intersection over union: A metric and a loss for bounding box regression.
\newblock In \emph{Proceedings of the IEEE/CVF conference on computer vision and pattern recognition}, pages 658--666, 2019.

\bibitem[Roddick et~al.(2018)Roddick, Kendall, and Cipolla]{roddick2018orthographic}
Thomas Roddick, Alex Kendall, and Roberto Cipolla.
\newblock Orthographic feature transform for monocular 3d object detection.
\newblock \emph{arXiv preprint arXiv:1811.08188}, 2018.

\bibitem[Rombach et~al.(2022)Rombach, Blattmann, Lorenz, Esser, and Ommer]{rombach2022high}
Robin Rombach, Andreas Blattmann, Dominik Lorenz, Patrick Esser, and Bj{\"o}rn Ommer.
\newblock High-resolution image synthesis with latent diffusion models.
\newblock In \emph{Proceedings of the IEEE/CVF conference on computer vision and pattern recognition}, pages 10684--10695, 2022.

\bibitem[Saharia et~al.(2022)Saharia, Chan, Saxena, Li, Whang, Denton, Ghasemipour, Gontijo~Lopes, Karagol~Ayan, Salimans, et~al.]{saharia2022photorealistic}
Chitwan Saharia, William Chan, Saurabh Saxena, Lala Li, Jay Whang, Emily~L Denton, Kamyar Ghasemipour, Raphael Gontijo~Lopes, Burcu Karagol~Ayan, Tim Salimans, et~al.
\newblock Photorealistic text-to-image diffusion models with deep language understanding.
\newblock \emph{Advances in Neural Information Processing Systems}, 35:\penalty0 36479--36494, 2022.

\bibitem[Schreier et~al.(2023)Schreier, Renz, Geiger, and Chitta]{schreier2023offline}
Tim Schreier, Katrin Renz, Andreas Geiger, and Kashyap Chitta.
\newblock On offline evaluation of 3d object detection for autonomous driving.
\newblock In \emph{Proceedings of the IEEE/CVF International Conference on Computer Vision}, pages 4084--4089, 2023.

\bibitem[Song et~al.(2020)Song, Meng, and Ermon]{song2020denoising}
Jiaming Song, Chenlin Meng, and Stefano Ermon.
\newblock Denoising diffusion implicit models.
\newblock \emph{arXiv preprint arXiv:2010.02502}, 2020.

\bibitem[Tur et~al.(2023)Tur, Dall’Asen, Beyan, and Ricci]{tur2023exploring}
Anil~Osman Tur, Nicola Dall’Asen, Cigdem Beyan, and Elisa Ricci.
\newblock Exploring diffusion models for unsupervised video anomaly detection.
\newblock In \emph{2023 IEEE International Conference on Image Processing (ICIP)}, pages 2540--2544. IEEE, 2023.

\bibitem[Vaswani et~al.(2017)Vaswani, Shazeer, Parmar, Uszkoreit, Jones, Gomez, Kaiser, and Polosukhin]{vaswani2017attention}
Ashish Vaswani, Noam Shazeer, Niki Parmar, Jakob Uszkoreit, Llion Jones, Aidan~N Gomez, {\L}ukasz Kaiser, and Illia Polosukhin.
\newblock Attention is all you need.
\newblock \emph{Advances in neural information processing systems}, 30, 2017.

\bibitem[Wang et~al.(2023{\natexlab{a}})Wang, Rupprecht, and Novotny]{wang2023posediffusion}
Jianyuan Wang, Christian Rupprecht, and David Novotny.
\newblock Posediffusion: Solving pose estimation via diffusion-aided bundle adjustment.
\newblock In \emph{Proceedings of the IEEE/CVF International Conference on Computer Vision}, pages 9773--9783, 2023{\natexlab{a}}.

\bibitem[Wang et~al.(2021)Wang, Zhu, Pang, and Lin]{wang2021fcos3d}
Tai Wang, Xinge Zhu, Jiangmiao Pang, and Dahua Lin.
\newblock Fcos3d: Fully convolutional one-stage monocular 3d object detection.
\newblock In \emph{Proceedings of the IEEE/CVF International Conference on Computer Vision}, pages 913--922, 2021.

\bibitem[Wang et~al.(2022{\natexlab{a}})Wang, Xinge, Pang, and Lin]{wang2022probabilistic}
Tai Wang, ZHU Xinge, Jiangmiao Pang, and Dahua Lin.
\newblock Probabilistic and geometric depth: Detecting objects in perspective.
\newblock In \emph{Conference on Robot Learning}, pages 1475--1485. PMLR, 2022{\natexlab{a}}.

\bibitem[Wang et~al.(2022{\natexlab{b}})Wang, Guizilini, Zhang, Wang, Zhao, and Solomon]{wang2022detr3d}
Yue Wang, Vitor~Campagnolo Guizilini, Tianyuan Zhang, Yilun Wang, Hang Zhao, and Justin Solomon.
\newblock Detr3d: 3d object detection from multi-view images via 3d-to-2d queries.
\newblock In \emph{Conference on Robot Learning}, pages 180--191. PMLR, 2022{\natexlab{b}}.

\bibitem[Wang et~al.(2023{\natexlab{b}})Wang, Yue, Hua, and Zhang]{wang2023random}
Yanghao Wang, Zhongqi Yue, Xian-Sheng Hua, and Hanwang Zhang.
\newblock Random boxes are open-world object detectors.
\newblock In \emph{Proceedings of the IEEE/CVF International Conference on Computer Vision}, pages 6233--6243, 2023{\natexlab{b}}.

\bibitem[Wu et~al.(2020)Wu, Chen, and Metaxas]{wu2020motionnet}
Pengxiang Wu, Siheng Chen, and Dimitris~N Metaxas.
\newblock Motionnet: Joint perception and motion prediction for autonomous driving based on bird's eye view maps.
\newblock In \emph{Proceedings of the IEEE/CVF conference on computer vision and pattern recognition}, pages 11385--11395, 2020.

\bibitem[Xie et~al.(2021)Xie, Cheng, Wang, Yao, and Han]{xie2021oriented}
Xingxing Xie, Gong Cheng, Jiabao Wang, Xiwen Yao, and Junwei Han.
\newblock Oriented r-cnn for object detection.
\newblock In \emph{Proceedings of the IEEE/CVF international conference on computer vision}, pages 3520--3529, 2021.

\bibitem[Yang et~al.(2023)Yang, Chen, Tian, Tao, Zhu, Zhang, Huang, Li, Qiao, Lu, et~al.]{yang2023bevformer}
Chenyu Yang, Yuntao Chen, Hao Tian, Chenxin Tao, Xizhou Zhu, Zhaoxiang Zhang, Gao Huang, Hongyang Li, Yu Qiao, Lewei Lu, et~al.
\newblock Bevformer v2: Adapting modern image backbones to bird's-eye-view recognition via perspective supervision.
\newblock In \emph{Proceedings of the IEEE/CVF Conference on Computer Vision and Pattern Recognition}, pages 17830--17839, 2023.

\bibitem[Ye et~al.(2023)Ye, Ke, Li, Tai, Tang, Danelljan, and Yu]{ye2023cascade}
Mingqiao Ye, Lei Ke, Siyuan Li, Yu-Wing Tai, Chi-Keung Tang, Martin Danelljan, and Fisher Yu.
\newblock Cascade-detr: Delving into high-quality universal object detection.
\newblock In \emph{Proceedings of the IEEE/CVF International Conference on Computer Vision}, pages 6704--6714, 2023.

\bibitem[Zhang et~al.(2022{\natexlab{a}})Zhang, Li, Liu, Zhang, Su, Zhu, Ni, and Shum]{zhang2022dino}
Hao Zhang, Feng Li, Shilong Liu, Lei Zhang, Hang Su, Jun Zhu, Lionel~M Ni, and Heung-Yeung Shum.
\newblock Dino: Detr with improved denoising anchor boxes for end-to-end object detection.
\newblock \emph{arXiv preprint arXiv:2203.03605}, 2022{\natexlab{a}}.

\bibitem[Zhang et~al.(2023)Zhang, Xiong, Yang, Casas, Hu, and Urtasun]{zhang2023learning}
Lunjun Zhang, Yuwen Xiong, Ze Yang, Sergio Casas, Rui Hu, and Raquel Urtasun.
\newblock Learning unsupervised world models for autonomous driving via discrete diffusion.
\newblock \emph{arXiv preprint arXiv:2311.01017}, 2023.

\bibitem[Zhang et~al.(2022{\natexlab{b}})Zhang, Zhu, Zheng, Huang, Huang, Zhou, and Lu]{zhang2022beverse}
Yunpeng Zhang, Zheng Zhu, Wenzhao Zheng, Junjie Huang, Guan Huang, Jie Zhou, and Jiwen Lu.
\newblock Beverse: Unified perception and prediction in birds-eye-view for vision-centric autonomous driving.
\newblock \emph{arXiv preprint arXiv:2205.09743}, 2022{\natexlab{b}}.

\bibitem[Zhao et~al.(2023)Zhao, Ye, Wu, Shen, and Wan]{zhao2023generative}
Yuzhong Zhao, Qixiang Ye, Weijia Wu, Chunhua Shen, and Fang Wan.
\newblock Generative prompt model for weakly supervised object localization.
\newblock In \emph{Proceedings of the IEEE/CVF International Conference on Computer Vision}, pages 6351--6361, 2023.

\bibitem[Zhu et~al.(2020)Zhu, Su, Lu, Li, Wang, and Dai]{zhu2020deformable}
Xizhou Zhu, Weijie Su, Lewei Lu, Bin Li, Xiaogang Wang, and Jifeng Dai.
\newblock Deformable detr: Deformable transformers for end-to-end object detection.
\newblock \emph{arXiv preprint arXiv:2010.04159}, 2020.

\bibitem[Zou et~al.(2023)Zou, Zhu, Ye, and Wang]{zou2023diffbev}
Jiayu Zou, Zheng Zhu, Yun Ye, and Xingang Wang.
\newblock Diffbev: Conditional diffusion model for bird's eye view perception.
\newblock \emph{arXiv preprint arXiv:2303.08333}, 2023.

\end{thebibliography}
}

\clearpage
\setcounter{page}{1}
\maketitlesupplementary

\begin{appendices}

In this document we provide additional reasoning, results, and comprehensive details supporting our method.

\section{Label Ambiguity}
\label{supp: label_ambiguity}

\begin{figure}[htbp]
    \centering
    \includegraphics[width=\linewidth]{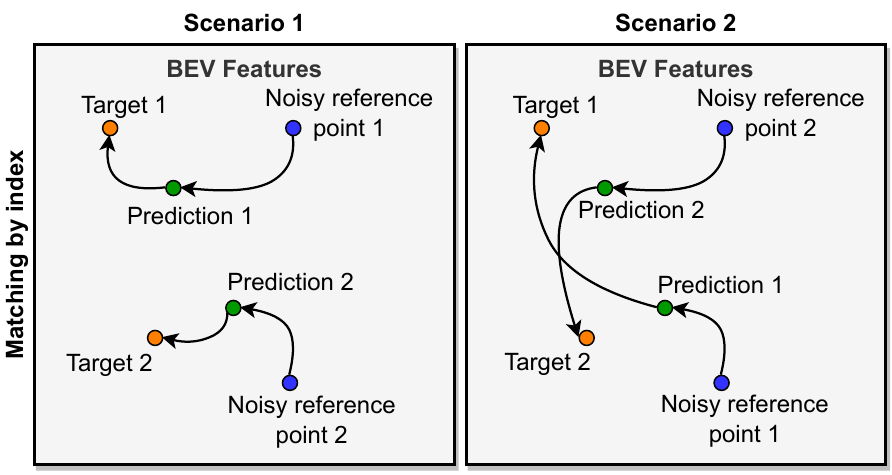}
    \caption{\textbf{Label ambiguity when matching by index}. Permuting the initially sampled reference points causes the matching to change which makes training unstable.}
    \label{fig: ambiguity_by_index}
\end{figure}

Label ambiguity arises when the matching between predictions and targets depends on the sampling of the references. Figure \ref{fig: label_ambiguity} showed an example where we match using the total distance. In Figure \ref{fig: ambiguity_by_index} we provide also a case where the matching is by index and label ambiguity is present. We assume there are $N$ targets $\mathbf{y}_1, ..., \mathbf{y}_N$ and $N$ predictions $\mathbf{b}_1, ..., \mathbf{b}_N$, and predictions $\mathbf{b}_i$ is matched with target $\mathbf{y}_i$.

When matching by index, suppose reference $\mathbf{r}_i$ has BEV coordinates $(x, y)$. The model either looks up the BEV features at $(x, y)$ or interpolates the queries at $(x, y)$ and produces a prediction which is matched to $\mathbf{y}_i$. However, if $\mathbf{r}_j$ is sampled at location $(x, y)$ the prediction will be the same, but the target will be $\mathbf{y}_j$. This confuses the model because the same features at $(x, y)$ can have different targets.

To assess the impact that ambiguous targets may have on the results, we study a simple toy task. We fix a single random image $\mathbf{I} \in \mathbb{R}^{C, H, W}$ and construct a network with the following forward pass:

\begin{itemize}
    \item First, the image is passed through two convolution layers that keep the output size the same as the initial size.
    \item Then, we look up, i.e. interpolate, the features at a number of reference points, provided as an additional input.
    \item The resulting features are processed by two linear layers after which new 2D locations are returned.
\end{itemize}

Thus, our network takes in an image and some reference points, and returns new 2D locations as output. The loss function is the simple $\ell_1$ loss between the predictions and a number of fixed targets. Our experiments suggest that when we sample the input reference points randomly, and there are only a few of them, the resulting label ambiguity is \emph{limiting} and prevents the network from overfitting, even on a single image. This is the case when there are fewer or equal random references than the number of targets. In Figure \ref{fig: label_ambiguity} we have 10 targets. With only 10 random references, the model does not converge, even if we let it run indefinitely.

With more references than targets, the model does manage to converge, with the convergence speed depending on the number of references. This is because only some predictions are used in the loss function, which makes predictions more localized to where their reference points start from. This behaviour does not exist if the references are fixed across training iterations, in which case the model always converges to a loss of zero, irrespective of whether it finds all targets or only some of them.

In a deterministic setup more references only speed up training. But when they are random, it becomes \emph{necessary} to have many of them in order to converge.

\section{Implementation}
\label{supp: implementation}
Here we provide additional information about the implementation and the experiments. Table \ref{table: hyperparameters} contains the training hyperparameters, while Algorithms \ref{algo: pseudocode_train} and \ref{algo: pseudocode_test} provide PyTorch-like pseudocodes for the training and testing logic. Most function names are borrowed from \cite{chen2023diffusion}.

\begin{table}[bp]
    \centering
    \small
    \begin{tabular}{ l l } \toprule[1.5pt]
         \textbf{Setting} &  \textbf{Value} \\ \midrule[1pt]
        Num. learnable object queries & 900 \\
        Num. references at test time & Variable \\
        BEV size, $(H, W, C)$ & $(200, 200, 256)$ \\
        Optimizer & AdamW \\
        Signal-to-noise ratio & 2 \\
        DDIM steps \cite{song2020denoising} & 3 \\
        NMS discard threshold & 0.1 \\
        Min. confidence threshold & 0.02 \\
        Radial suppression threshold & 0.5 \\
        Query formation & Interpolation \\
        Matching type & \emph{Simple} many-to-one \\
        BEV encoder & BEVFormer-Base \cite{li2022bevformer} \\
        Decoder type & DeformableDETR \cite{zhu2020deformable} \\
        Training epochs & 24 \\
        LR schedule & Cosine, $2 \times 10^{-4} \rightarrow 10^{-6}$ \\
        Clip gradient 2-norm & 35 \\
        Batch size & 1 \\
        \bottomrule[1.5pt]
    \end{tabular}
    \caption{\textbf{Particle-DETR training hyperparameters}. Additional hyperparameters follow BEVFormer-Base \cite{li2022bevformer}.}
    \label{table: hyperparameters}
\end{table}

\begin{algorithm}[tbp]
\SetAlgoLined
    \footnotesize
    \PyComment{Inputs} \\
    \Indp
        \PyComment{F: BEV features, (C, H, W)} \\
        \PyComment{fixed\_queries: queries over which to interpolate, (N, C)} \\
        \PyComment{GTs: ground-truth boxes, (N, C)} \\ 
        \PyComment{scale: the signal-to-noise ratio} \\
    \Indm

    \PyCode{} \\
    
    \PyComment{Extract object centers in BEV} \\
    \PyCode{GT\_centers = GTs[..., :2]} \\
    \PyCode{} \\
    
    \PyComment{Pad references up to a desired number} \\
    \PyCode{ref\_points = pad\_refs(GT\_centers)} \\
    \PyCode{} \\
    
    \PyComment{Scale and apply diffusion} \\
    \PyCode{ref\_points = (2 * ref\_points - 1) * scale} \\
    \PyCode{diff\_time = randint(0, T)} \\
    \PyCode{eps = normal(mean=0, std=1)} \\
    \PyCode{ref\_points = sqrt(alpha\_cumprod(diff\_time)) * ref\_points + sqrt(1 - alpha\_cumprod(diff\_time)) * eps} \\
    \PyCode{} \\

    \PyComment{Interpolate the queries} \\
    \PyCode{queries = grid\_sample(fixed\_queries, ref\_points)} \\
    \PyCode{keys = F} \\
    \PyCode{values = F} \\ 
    \PyCode{} \\

    \PyComment{Call the decoder and compute loss} \\
    \PyCode{pred\_boxes = decoder(queries, keys, values, ref\_points, diff\_time)} \\
    \PyCode{loss = set\_prediction\_loss(GTs, pred\_boxes)} \\

\caption{Particle-DETR Training}
\label{algo: pseudocode_train}
\end{algorithm}

For the implementation, our codebase is based on that of BEVFormer \cite{li2022bevformer}. We train all models for 24 epochs on the NuScenes dataset \cite{caesar2020nuscenes} on 8 NVidia A100 GPUs, while the evaluation is always performed on a single GPU. At both training and test time the batch size is set to 1.

\textbf{Diffusion box updates.} The original DiffusionDet \cite{chen2023diffusion} only works with axis-aligned boxes. For the baseline, we re-implement and modify it to use rotated boxes. Inspired by \cite{xie2021oriented}, each stage of the decoder outputs the elements $(\delta c_x, \delta c_y, c_z, \delta w, \delta h, l, \delta \theta, v_x, v_y)$, which are applied to the input boxes $(c_x, c_y, w, h, \theta)$ to produce the updated boxes at the current stage $(c_x', c_y', w', h', \theta')$ as follows:
\begin{align*}
    \bar{w} &= w \cos \theta + h \sin \theta \\
    \bar{h} &= w \sin \theta + h \cos \theta \\
    c_x' &= \bar{w} \delta c_x  + c_x \\
    c_y' &= \bar{h} \delta c_y + c_y \\
    w' &= \exp(\delta w) w \\
    h' &= \exp(\delta h) h \\
    \theta' &= \theta + \delta \theta.
\end{align*}

\textbf{GIoU loss}. Unlike the original DiffusionDet \cite{chen2023diffusion}, in our implementation we do not use the GIoU loss \cite{rezatofighi2019generalized} because its implementation for rotated boxes with supported backpropagation is non-trivial. We leave the investigation of how similar metrics can be used as losses for future work.

\textbf{SimOTA.} When evaluating the simOTA matching strategy \cite{ge2021yolox} we remove the cost matrix masking which is present in DiffusionDet \cite{chen2023diffusion}. There, additional cost is added to those predictions whose center does not fall close enough to or within a target box, which effectively prevents these predictions from ever being matched. In our setup, this masking introduced a large amount of instability. Hence, we utilize simOTA using the raw cost matrix, where all predictions are considered as potential matchings to all targets.

\begin{algorithm}[tbp]
\SetAlgoLined
    \footnotesize
    \PyComment{Inputs} \\
    \Indp
        \PyComment{F: BEV features, (C, H, W)} \\
        \PyComment{fixed\_queries: queries over which to interpolate, (N, C)} \\
    \Indm

    \PyCode{} \\
    
    \PyComment{Prepare random references} \\
    \PyCode{eps = normal(mean=0, std=1)} \\
    \PyCode{ref\_points = normalize(eps)} \\
    \PyComment{DDIM times [(T-1, T-2), ..., (0, -1)]} \\
    \PyCode{times = reversed(linespace(-1, T, steps))} \\
    \PyCode{time\_pairs = list(zip(times[:-1], times[1:])} \\
    
    \PyCode{} \\
    \PyComment{Interpolate the queries} \\
    \PyCode{queries = grid\_sample(fixed\_queries, ref\_points)} \\
    \PyCode{keys = F} \\
    \PyCode{values = F} \\ 
    \PyCode{} \\
    
    \PyCode{all\_preds = []} \\
    \PyCode{for (t\_now, t\_next) in time\_pairs:} \\

    \Indp
        \PyCode{pred\_boxes, queries = decoder(queries, keys, values, ref\_points, t\_now)} \\
        \PyCode{all\_preds.append(pred\_boxes)} \\
        \PyCode{ref\_points = ddim\_step(ref\_points, pred\_boxes, t\_now, t\_next)} \\
        \PyCode{ref\_points = ref\_renewal(ref\_points)} \\
    \Indm

    \PyCode{} \\
    \PyComment{Filter predictions} \\
    \PyCode{preds = nms(all\_preds)} \\
    \PyCode{preds = radial\_suppression(preds)} \\
\caption{Particle-DETR Inference}
\label{algo: pseudocode_test}
\end{algorithm}

\section{Additional Experiments}
\label{supp: additional_experiments}
Here we provide additional experimental results. All results and plots are from the NuScenes \cite{caesar2020nuscenes} validation dataset, unless otherwise noted.

\textbf{Deterministic and random references.} Our Particle-DETR provides rich opportunities to tweak the test-time performance after training. Once the model is trained, one can freely change the hyparparameters governing the inference behaviour. First, we assess how the joint training with two sets of references, fixed and random, affects performance. Once the model from this joint setup is trained, we can evaluate with both sets of references or with either one of them. Table \ref{table: average_metrics_bevformer_enh_with_random_queries} shows the performance when evaluating with both sets. The performance is statistically-significant and outperforms BEVFormer on all metrics.

\begin{table}[tbp]
    \centering
    \small
    \begin{tabular}[width=\textwidth]{ l | l | l l } \toprule[1.5pt]
          DDIM steps & Diff. queries & mAP $\uparrow$ & NDS $\uparrow$  \\
         \midrule[1pt]
         \multirow{6}{*}{1} & 300 & 0.4192 & 0.5301  \\
         &                                     500 & 0.4192 & 0.5302  \\ 
         &                                      700 & 0.4202 & 0.5306  \\ 
         &                                      900 & 0.4211 & 0.5310 \\ 
         &                                      1200 & 0.4208 & 0.5304 \\ 
         &                                      1500 & 0.4208 & 0.5302  \\ 
         \hline
         \multirow{6}{*}{3} & 300 & 0.4109 & 0.5245  \\
         &                                     500 & 0.4155 & 0.5275  \\ 
         &                                      700 & 0.4171 & 0.5285  \\ 
         &                                      900 & 0.4188 & 0.5289 \\ 
         &                                      1200 & 0.4184 & 0.5285  \\ 
         &                                      1500 & 0.4192 & 0.5290 \\ 
         \hline
         \multirow{6}{*}{5} & 300 & 0.4164 & 0.5275  \\
         &                                     500 & 0.4190 & 0.5288  \\ 
         &                                      700 & 0.4201 & 0.5295 \\ 
         &                                      900 & 0.4198 & 0.5290 \\ 
         &                                      1200 & 0.4196 & 0.5292  \\ 
         &                                      1500 & 0.4197 & 0.5293  \\ 
         \hline
         \multirow{6}{*}{7} & 300 & 0.4172 & 0.5281  \\
         &                                     500 & 0.4194 & 0.5289  \\ 
         &                                      700 & 0.4203 & 0.5297  \\ 
         &                                      900 & 0.4199 & 0.5290  \\ 
         &                                      1200 & 0.4201 & 0.5292  \\ 
         &                                      1500 & 0.4189 & 0.5287  \\ 
         \hline
         \multirow{6}{*}{9} & 300 & 0.4179 & 0.5280  \\
         &                                     500 & 0.4192 & 0.5290  \\ 
         &                                      700 & 0.4198 & 0.5294  \\ 
         &                                      900 & 0.4197 & 0.5291  \\ 
         &                                      1200 & 0.4186 & 0.5287  \\ 
         &                                      1500 & 0.4184 & 0.5284  \\ 
        \bottomrule[1.5pt]
    \end{tabular}
    \captionsetup{belowskip=-0.4cm}
    \caption{\textbf{Joint effects of DDIM steps and diffusion object queries when using both diffusion and fixed queries.} With both query sets, adding more random queries or DDIM steps does not have any noticeable effect. In practice, one could use just a single DDIM step with a variable number of queries.}
    \label{table: ddim_and_queries_both_sets}
\end{table}

We analyze how each reference set contributes the results. To do this, we look at the self-attention values throughout the decoder layers. When evaluating with both query sets, fixed queries from the first decoder layer spend about 94\% of their attention on other fixed queries, whereas the diffusion queries, associated with the random references, spend 82.5\% of their attention on the fixed queries. This imbalance is rectified in the subsequent decoder layers. Particularly, in the last decoder layer both query sets spend about 50\% of their attention onto the other query set.

\begin{table}[tbp]
    \centering
    \small
    \begin{tabular}[width=\textwidth]{ l | l | l l } \toprule[1.5pt]
          DDIM steps & Diff. queries & mAP $\uparrow$ & NDS $\uparrow$  \\
         \midrule[1pt]
         \multirow{6}{*}{1} & 300 & 0.3540 & 0.4936  \\
         &                                     500 & 0.3907 & 0.5140 \\ 
         &                                      700 & 0.4046 & 0.5208  \\ 
         &                                      900 & 0.4105 & 0.5242  \\ 
         &                                      1200 & 0.4143 & 0.5261  \\ 
         &                                      1500 & 0.4162 & 0.5273  \\ 
         \hline
         \multirow{6}{*}{3} & 300 & 0.3755 & 0.5038  \\
         &                                     500 & 0.3944 & 0.5141  \\ 
         &                                      700 & 0.4032 & 0.5191 \\ 
         &                                      900 & 0.4076 & 0.5208  \\ 
         &                                      1200 & 0.4114 & 0.5232 \\ 
         &                                      1500 & 0.4127 & 0.5242  \\ 
         \hline
         \multirow{6}{*}{5} & 300 & 0.3930 & 0.5122  \\
         &                                     500 & 0.4062 & 0.5201  \\ 
         &                                      700 & 0.4111 & 0.5234  \\ 
         &                                      900 & 0.4133 & 0.5250  \\ 
         &                                      1200 & 0.4144 & 0.5253  \\ 
         &                                      1500 & 0.4154 & 0.5263  \\ 
         \hline
         \multirow{6}{*}{7} & 300 & 0.3999 & 0.5164  \\
         &                                     500 & 0.4096 & 0.5224  \\ 
         &                                      700 & 0.4123 & 0.5241 \\ 
         &                                      900 & 0.4144 & 0.5257 \\ 
         &                                      1200 & 0.4148 & 0.5259  \\ 
         &                                      1500 & 0.4154  & 0.5265  \\ 
         \hline
         \multirow{6}{*}{9} & 300 & 0.4019 & 0.5175  \\
         &                                     500 & 0.4110 & 0.5237  \\ 
         &                                      700 & 0.4134 & 0.5249  \\ 
         &                                      900 & 0.4149 & 0.5259  \\ 
         &                                      1200 & 0.4148 & 0.5261  \\ 
         &                                      1500 & 0.4147 & 0.5258  \\ 
        \bottomrule[1.5pt]
    \end{tabular}
    \captionsetup{belowskip=-0.4cm}
    \caption{\textbf{The joint effects of DDIM steps and diffusion object queries when utilizing diffusion queries only.} Here, we explore whether adding more DDIM steps or more diffusion queries improves performance. When using only diffusion queries, more DDIM steps and more queries improve performance.}
    \label{table: ddim_and_queries_diff_only}
\end{table}

\begin{table}[b]
    \centering
    \small
    \begin{tabular}{ l c c } \toprule[1.5pt]
         \textbf{Metric} &  BEVFormer-Base & BEVFormer-Enh (ours) \\
         \midrule[1pt]
         NDS $\uparrow$ & 0.5168 & \textbf{0.5287}  (0.0003)\\
         mAP $\uparrow$  & 0.4154  & \textbf{0.4184} (0.0006) \\ 
         mATE $\downarrow$ & 0.6715 & \textbf{0.6386} (0.0009) \\
         mASE $\downarrow$ & 0.2738 & \textbf{0.2686} (0.0002) \\
         mAOE $\downarrow$ & 0.3691 & \textbf{0.3362} (0.0009)\\
         mAVE $\downarrow$ & 0.4179 & \textbf{0.3688} (0.0013)\\
        mAAE $\downarrow$ & 0.1981 & \textbf{0.1931} (0.0007)\\ 
        \bottomrule[1.5pt]
    \end{tabular}
    \captionsetup{belowskip=-0.4cm}
    \caption{\textbf{Performance of BEVFormer-Enh}. We compare BEVFormer-Enh, evaluated with both static and random references, to the fully-deterministic BEVFormer. For our method, we repeat the evaluation 10 times and report the mean values. The standard deviations are shown in parentheses.}
    \label{table: average_metrics_bevformer_enh_with_random_queries}
\end{table}

When evaluating the final model with both static and random references we find that increasing the number of DDIM steps or the number of random queries does not have a significant effect, presumably because the queries corresponding to the static references are more important. This is highlighted in Table \ref{table: ddim_and_queries_both_sets}. The mAP and NDS results are slightly higher than if we are using the static queries only, as in Table \ref{table: average_metrics}, showing that the additional queries corresponding to random references do increase performance.

\begin{table}[tbp]
    \centering
    \small
    \begin{tabular}[width=\textwidth]{ l | l | l l } \toprule[1.5pt]
          DDIM steps & Diff. queries & mAP $\uparrow$ & NDS $\uparrow$  \\
         \midrule[1pt]
         \multirow{6}{*}{1} & 300 & 0.3540 & 0.4931  \\
         &                                     500 & 0.3899 & 0.5125  \\ 
         &                                      700 & 0.4042 & 0.5202  \\ 
         &                                      900 & 0.4108 & 0.5239 \\ 
         &                                      1200 & 0.4142 & 0.5260 \\ 
         &                                      1500 & 0.4164 & 0.5272 \\ 
         \hline
         \multirow{6}{*}{3} & 300 & 0.3746 & 0.5031  \\
         &                                     500 & 0.3952 & 0.5143  \\ 
         &                                      700 & 0.4034 & 0.5187  \\ 
         &                                      900 & 0.4080 & 0.5214 \\ 
         &                                      1200 & 0.4113 & 0.5238 \\ 
         &                                      1500 & 0.4139 & 0.5250  \\ 
         \hline
         \multirow{6}{*}{5} & 300 & 0.3940 & 0.5128  \\
         &                                     500 & 0.4070 & 0.5210  \\ 
         &                                      700 & 0.4124 & 0.5244  \\ 
         &                                      900 & 0.4146 & 0.5257  \\ 
         &                                      1200 & 0.4159 & 0.5267  \\ 
         &                                      1500 & 0.4169 & 0.5269  \\ 
         \hline
         \multirow{6}{*}{7} & 300 & 0.3999 & 0.5162  \\
         &                                     500 & 0.4104 & 0.5230  \\ 
         &                                      700 & 0.4133 & 0.5244  \\ 
         &                                      900 & 0.4154 & 0.5263  \\ 
         &                                      1200 & 0.4162 & 0.5267  \\ 
         &                                      1500 & 0.4165  & 0.5270 \\ 
         \hline
         \multirow{6}{*}{9} & 300 & 0.4031 & 0.5175  \\
         &                                     500 & 0.4120 & 0.5240 \\ 
         &                                      700 & 0.4142 & 0.5248  \\ 
         &                                      900 & 0.4156 & 0.5263  \\ 
         &                                      1200 & 0.4161 & 0.5270 \\ 
         &                                      1500 & 0.4165 & 0.5272  \\ 
        \bottomrule[1.5pt]
    \end{tabular}
    \captionsetup{belowskip=-0.4cm}
    \caption{\textbf{The joint effects of DDIM steps and diffusion object queries when utilizing radial suppression.} When using only diffusion queries, more DDIM steps and more queries improve performance considerably.}
    \label{table: ddim_and_queries_diff_only_with_radial_suppression}
\end{table}

What happens if we use only the random references? In that case the mAP and NDS metrics are naturally lower, because the proposed corrections should all be relative to the current reference location, which is random. There is a clear trade-off between the usage of additional DDIM steps or additional references, and performance. Tables \ref{table: ddim_and_queries_diff_only} and \ref{table: ddim_and_queries_diff_only_with_radial_suppression} show this with and without radial suppression. We highlight that given enough random references the Particle-DETR does manage to beat BEVFormer on mAP. On NDS, it only takes 900 references and a single DDIM step to beat it.

\begin{table}[tbp]
    \centering
    \small
    \begin{tabular}[width=\textwidth]{ l | l | l | l l } \toprule[1.5pt]
          NMS threshold & Conf. threshold & mAP $\uparrow$ & NDS $\uparrow$   \\
         \midrule[1pt]
         \multirow{3}{*}{0.1} & 0.02 & 0.3845 & 0.5135  \\
         &                                      0.05 & 0.3818 & 0.5138  \\ 
         &                                      None & 0.3847 & 0.5132  \\ 
         \hline
         \multirow{3}{*}{0.5} & 0.02 & 0.3876 & 0.5129  \\
         &                                      0.05 & 0.3859 & 0.5142  \\ 
         &                                      None & 0.3876 & 0.5128  \\ 
         \hline
         \multirow{3}{*}{None} & 0.02 & 0.2264 & 0.4305  \\
         &                                      0.05 & 0.2261 & 0.4314 \\ 
         &                                      None & 0.2264 & 0.4303  \\ 
        \bottomrule[1.5pt]
    \end{tabular}
    \captionsetup{belowskip=-0.4cm}
    \caption{\textbf{Filtering using NMS and confidence thresholds}. We evaluate a Particle-DETR trained with SimOTA \cite{ge2021yolox} matching without radial suppression. We set the DDIM steps to $3$ and vary the NMS threshold and the score threshold. The results show that NMS is needed. Furthermore, using SimOTA matching does not result in better performance than \emph{simple} many-to-one matching.}
    \label{table: nms_and_score_threshold}
\end{table}

\begin{table}[bp]
    \centering
    \small
    \begin{tabular}[width=\textwidth]{ l l l} \toprule[1.5pt]
          Setting & mAP $\uparrow$ & NDS $\uparrow$ \\
         \midrule[1pt]
         $r = 0.5$ m & 0.4112 & 0.5234 \\
         $r = 0.75$ m & 0.4143 & 0.5257  \\
         $r = 1$ m & 0.4132 & 0.5257 \\
         $r = 1.25$ m & 0.4102 & 0.5242  \\
         $r = 1.5$ m & 0.4073 & 0.5226  \\
         $r = 1.75$ m & 0.4030 & 0.5204  \\
         $r = 2$ m & 0.3971 & 0.5172  \\
         $r = 3$ m & 0.3697 & 0.5014  \\
         $r = 4$ m & 0.3372 & 0.4806  \\
        \bottomrule[1.5pt]
    \end{tabular}
    \caption{\textbf{The effect of radial suppression}. Here, we set the NMS threshold to $0.5$, the DDIM steps to $3$, the score threshold to $0.02$ and only vary the radius of the radial suppression. The results suggest that a small radius is best because it keeps the averaging sufficiently localized. Larger radii cause predictions corresponding to different targets to be averaged, reducing detection accuracy.}
    \label{table: radial_suppression}
\end{table}

\textbf{Filtering.} We also provide additional justification for why filtering is needed. Since the many-to-one matching causes multiple predictions to be stacked on top of each other we found the usage of NMS necessary. The best threshold is $0.1$, which we also combine with confidence-based filtering, as shown in Table \ref{table: nms_and_score_threshold}.

Our radial suppression replaces a confident predicted box with the weighted average of the predictions within a ball neighborhood. We tune the radius of this neighborhood. For very small values no predictions are filtered. For large values predictions from multiple different objects are filtered. The optimal occurs around 0.5 meters, as shown in Table \ref{table: radial_suppression}. Data for the joint tuning of the radius and the NMS can be found in Table \ref{table: nms_and_radius}.

\textbf{Reference resampling}. Following DiffusionDet \cite{chen2023diffusion} we resample the search tokens between DDIM steps. We experimented with different strategies but found the basic one in \cite{chen2023diffusion} to work best, as shown in Table 
\ref{table: resampling_strategies}. Thus, between each DDIM step we simply resample the references corresponding to less confident predictions. Alternative strategies which we tested include resampling close to the confident predictions, resampling without applying the DDIM steps, or no resampling altogether.

\textbf{Model characteristics}. Our BEVFormer-Enh model has the same number of parameters, FLOPS, and FPS as BEVFormer \cite{li2022bevformer}. The Particle-DETR is similar but can use more compute depending on how many DDIM steps one runs. The compute in the query interpolation depends only on the number of reference points and not on the DDIM steps.

\begin{table}[htbp]
    \centering
    \small
    \begin{tabular}[width=\textwidth]{ l | l | l l } \toprule[1.5pt]
          \multicolumn{2}{c|}{Setting} & mAP $\uparrow$ & NDS $\uparrow$  \\
         \midrule[1pt]
         \multirow{5}{*}{NMS threshold = 0.1} & $r = 0.5$ m & \textbf{0.4188} & \textbf{0.5289}  \\
         &                                      $r = 0.75$ m & 0.4173 & 0.5283  \\ 
         &                                      $r = 1$ m & 0.4138 & 0.5265  \\ 
         &                                      $r = 1.25$ m & 0.4101 & 0.5246  \\ 
         &                                      $r = 1.5$ m & 0.4066 & 0.5228  \\ 
         \hline
         \multirow{5}{*}{NMS threshold = 0.5} & $r = 0.5$ m & 0.4112 & 0.5237  \\
         &                                      $r = 0.75$ m & \textbf{0.4130} & \textbf{0.5251}  \\ 
         &                                      $r = 1$ m & 0.4123 & 0.5250 \\ 
         &                                      $r = 1.25$ m & 0.4096 & 0.5234  \\ 
         &                                      $r = 1.5$ m & 0.4067 & 0.5222  \\ 
        \bottomrule[1.5pt]
    \end{tabular}
    \captionsetup{belowskip=-0.4cm}
    \caption{\textbf{The joint effects of radial suppression and NMS}. Here, we set the DDIM steps to $3$ and the score threshold to $0.02$ and only vary the NMS threshold and the radius of the radial suppression. The best results are achieved with strong NMS suppression and a relatively small radius.}
    \label{table: nms_and_radius}
\end{table}

\begin{table}[htbp]
    \centering
    \small
    \begin{tabular}[width=\textwidth]{ l l l} \toprule[1.5pt]
          Resampling strategy & mAP $\uparrow$ & NDS $\uparrow$ \\
         \midrule[1pt]
         Standard normal resampling \cite{chen2023diffusion} & \textbf{0.4156} & \textbf{0.5264} \\
         Resample near predictions & 0.4144 & 0.5251  \\
         No DDIM, with resampling & 0.4146 & 0.5252 \\
         No DDIM, no resampling & 0.3959 & 0.5130  \\
        \bottomrule[1.5pt]
    \end{tabular}
    \captionsetup{belowskip=-0.4cm}
    \caption{\textbf{Different resampling strategies}. The simple strategy of pruning the low-confident references are replacing them with new random ones works best.}
    \label{table: resampling_strategies}
\end{table}

\begin{table}[htbp]
    \centering
    \small
    \begin{tabular}{ l c c } \toprule[1.5pt]
         \textbf{Metric} &  Particle-DETR & BEVFormer-Enh \\
         \midrule[1pt]
         NDS $\uparrow$ & 0.5283 & 0.5323  \\
         mAP $\uparrow$  & 0.4287  & 0.4326  \\ 
         mATE $\downarrow$ & 0.6011 & 0.5904  \\
         mASE $\downarrow$ & 0.2600 & 0.2595  \\
         mAOE $\downarrow$ & 0.4596 & 0.4504 \\
         mAVE $\downarrow$ & 0.4125 & 0.4109 \\
        mAAE $\downarrow$ & 0.1275 & 0.1287 \\ 
        \bottomrule[1.5pt]
    \end{tabular}
    \captionsetup{belowskip=-0.4cm}
    \caption{\textbf{Nuscenes \texttt{test} set metrics}. For the Particle-DETR we use 1500 queries and 1 DDIM step, with radial suppression radius set to 0.5 and NMS threshold set to 0.1.}
    \label{table: test_set_metrics}
\end{table}

\begin{figure}[tbp]
    \centering
    \includegraphics[width=\linewidth]{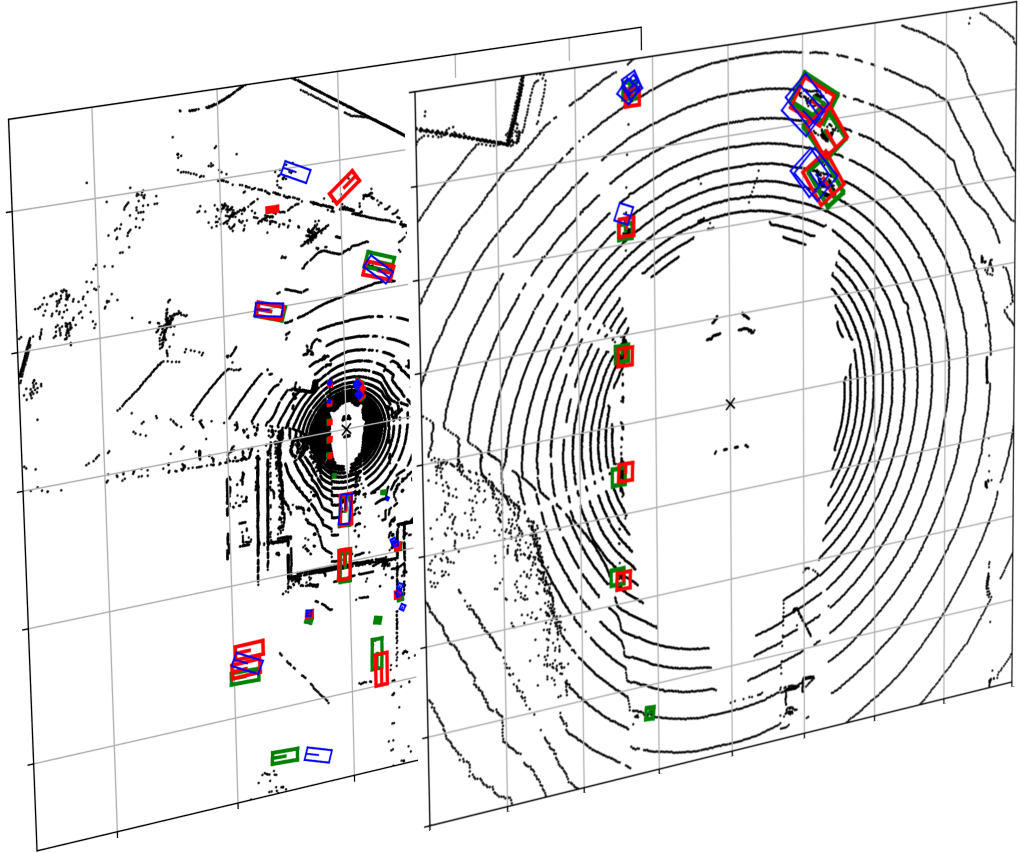}
    \captionsetup{belowskip=-0.4cm}
    \caption{\textbf{Sample predictions in BEV}. Green boxes are ground-truths, red are predicted by our Particle-DETR, and blue is predicted by BEVFormer. The right figure is a zoomed-in version of the left one, centered around the ego-vehicle.}
    \label{fig: comparison_3}
\end{figure}

\begin{figure}[b]
    \centering
    \includegraphics[width=\linewidth]{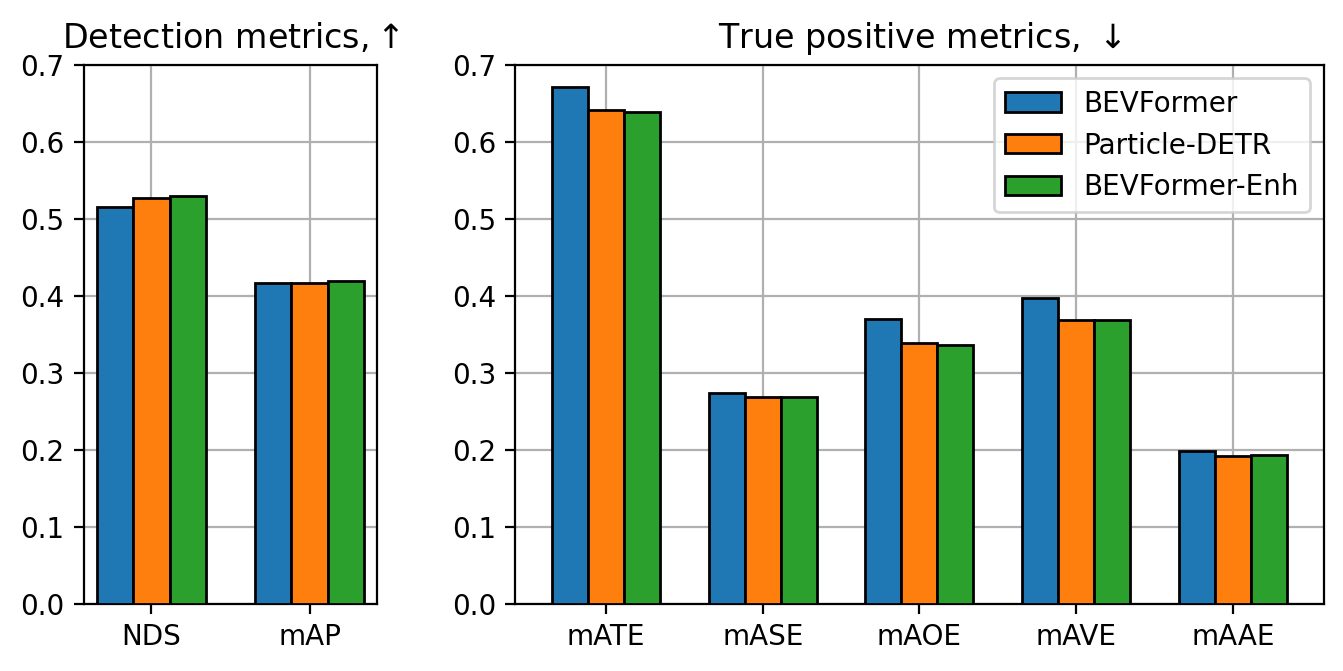}
    \caption{\textbf{Metrics comparison}. The improved NDS results from significantly lower orientation, translation, and velocity errors.}
    \label{fig: metrics_barchart}
\end{figure}

\begin{figure*}[tbp]
    \centering
    \includegraphics[width=\linewidth]{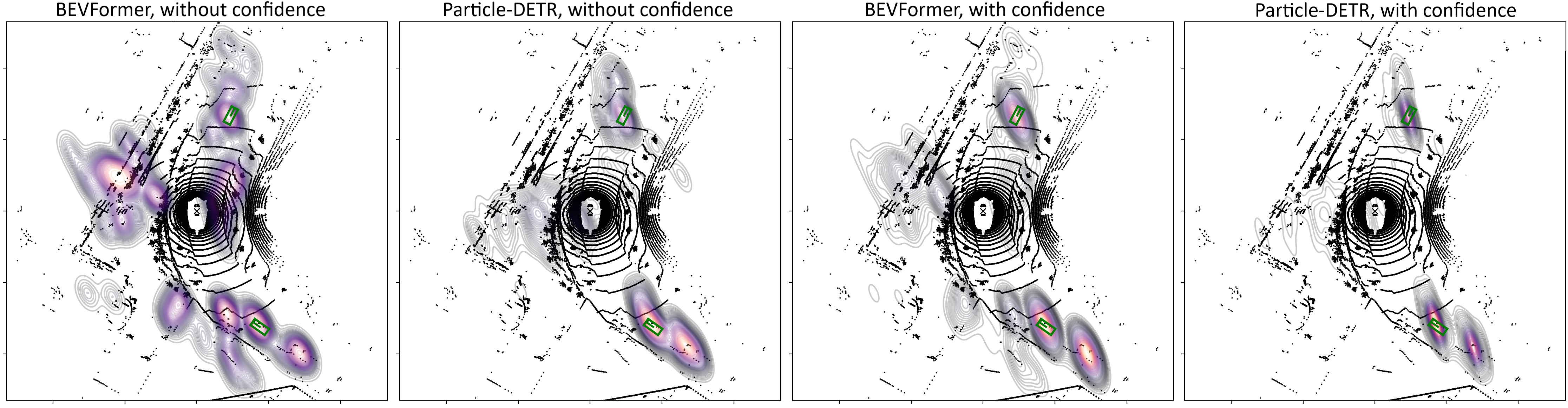}
    \captionsetup{belowskip=-0.4cm}
    \caption{\textbf{Uncertainty comparison}. Our Particle-DETR produces meaningful heatmaps due to its many-to-one attractive nature.}
    \label{fig: uncertainty_comparison}
\end{figure*}

\begin{figure}[b]
    \centering
    \includegraphics[width=\linewidth]{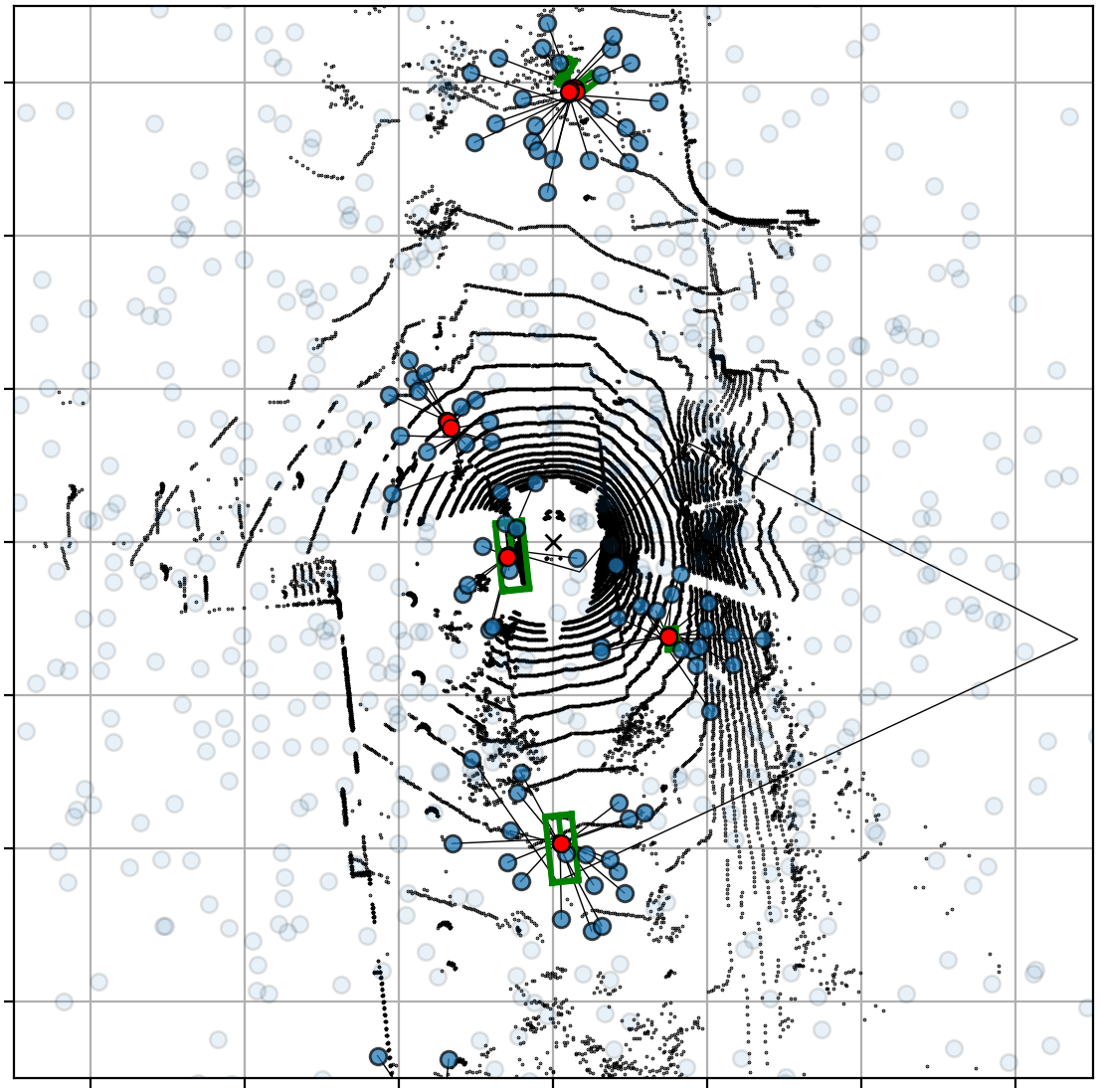}
    \caption{\textbf{Reference dynamics}. We plot the starting random references as faint blue circles. Predicted centers with sufficient confidence are shown in red. The starting references transformed to those predictions are in brighter blue. The black lines show how each reference has been modified through the six decoder stages. Better viewed zoomed.}
    \label{fig: prediction_dynamics}
\end{figure}

\begin{figure*}[tbp]
    \centering
    \includegraphics[width=\linewidth]{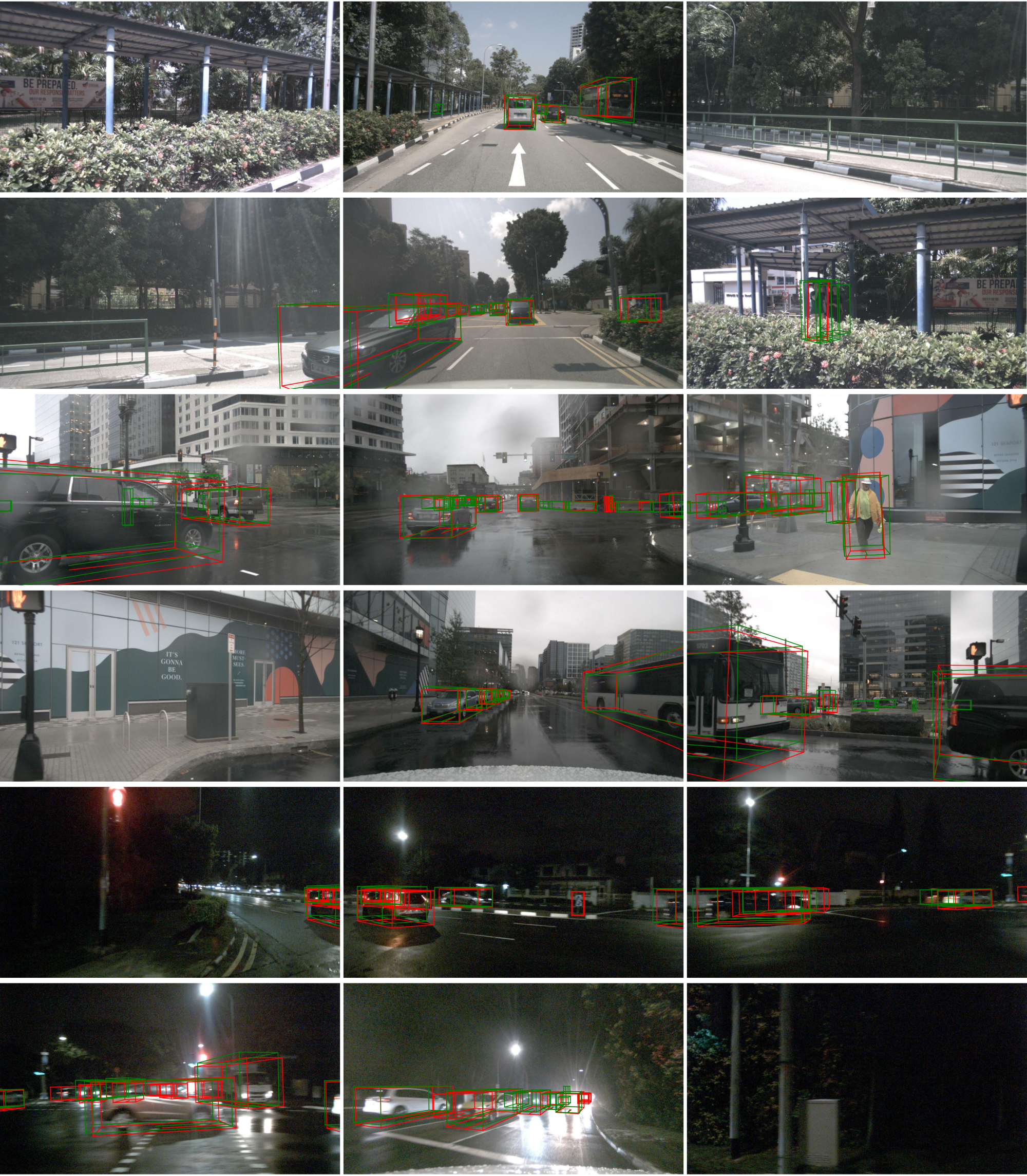}
    \caption{\textbf{Particle-DETR predictions projected onto the camera images}. We show three scenes, each with six cameras around the ego-vehicle. Green boxes are ground-truths, red are predicted by our Particle-DETR. We only show predictions with confidence greater than 0.2. The predictions are relatively accurate across diverse driving scenarios, including sunny, overcast, and night-time conditions.}
    \label{fig: preds_frontal}
\end{figure*}

\section{Qualitative Analysis}
Here we provide additional visualizations of the predictions. Figure \ref{fig: comparison_3} shows an example where our Particle-DETR, using only the diffusion queries, detects very small objects which are missed by BEVFormer. In fact, the AP at 0.5 meters for traffic cones is above 0.34, whereas that of BEVFormer is 0.28. Similar detection boosts can be observed also for cars (+1 AP point), bicycles (+2.4), motorcycles (+2.8), pedestrians (+1.8), and barriers (+9 AP points).

In general, the increased NDS results mostly from more accurate translation, orientation, and velocity. In Figure \ref{fig: preds_frontal} we show predictions from our Particle-DETR projected onto the camera images. We highlight diverse driving conditions including bright sunshine, rain - where raindrops create localized blur in the images, and nighttime - where pixel intensity noise due to the low exposure time is present. In all these cases our method produces reasonably accurate predictions, while being a fully generative model.

\textbf{Basins of attraction.} We visualize the transformation of the starting references in Figure \ref{fig: prediction_dynamics}. In general, each GT attracts the starting references around it. This \emph{gradient flow} is learned by the model and the many-to-one matching is a necessary condition for its existence. The basins of attraction are well localized and separated. We considered adding $\ell_1$ regularization between the predictions and the starting references to explicitly make predictions more localized but this was not needed as such a localization property seems to develop naturally from the training setup.

The BEV is patched together from multiple camera views. Logically, it is desirable to prohibit reference points starting from one side of the ego-vehicle to be refined to the other side because a view from one side does not provide information about the opposite view. The attractive nature of the Particle-DETR more or less satisfies this constraint.

The references which start in problematic BEV regions, such as behind walls or outside of the road, are pushed to the sides of the BEV as the features at those locations do not correspond to any visible scene elements. The confidence of the corresponding predictions is near zero.

\section{Additional Discussion}
Standard DETR models are fully-deterministic. To get an uncertainty estimate over the predictions, one usually needs to explicitly modify the model architecture, for example by adding additional outputs, which stand in for the standard deviation of the box location. Here, our generative Particle-DETR is advantageous, in that a rudimentary form of uncertainty may be readily available.

First, we consider the baseline BEVFormer and we plot heatmaps over the predicted box centers, computed using kernel density estimation, as shown in Figure \ref{fig: uncertainty_comparison}. The density (and color) at each predicted center in this plot is determined mainly by how close this point is to nearby predictions. The first heatmap shows the density over the predicted centers only, even without considering the confidence of each prediction. Since BEVFormer uses one-to-one matching, most of the predictions are quite spread apart and few of them are attracted to the same GT box. If we weight the predictions by their confidence, we get the third heatmap, which is more reasonable.

Now, we apply the same procedure to our Particle-DETR. In the second heatmap we plot the density of the predicted centers only, \emph{before applying NMS}. Due to the many-to-one matching during training, the predictions stack, which prevents the density from being too spread apart and keeps it relatively focused on the true objects. If we further consider the predicted confidence as a weight to each point, we get the fourth heatmap where the density is even better localized.

Thus, the dynamics of how predictions are formed themselves contain information. Though for object detection we only use a few predictions, for uncertainty estimation many can be useful. We leave it for future work to investigate how to reason more formally about this opportunity.

\end{appendices}

\end{document}